\documentclass[11pt,a4paper]{article}
\usepackage[hyperref]{emnlp2020}
\usepackage{times}
\usepackage{latexsym}

\usepackage{bm}
\usepackage{microtype}
\usepackage{hyperref}
\usepackage{url}
\usepackage{wrapfig,lipsum,booktabs}
\usepackage{xcolor}
\usepackage{colortbl}
\usepackage[normalem]{ulem}
\definecolor{myblue}{rgb}{0.54, 0.81, 0.94}
\definecolor{mygreen}{rgb}{0.64, 0.76, 0.68}
\definecolor{myyellow}{rgb}{0.98, 0.94, 0.75}
\definecolor{mygreen}{rgb}{0.68, 0.85, 0.9}
\definecolor{mypink}{rgb}{0.99, 0.87, 0.9}
\definecolor{myblue}{rgb}{0.82, 0.94, 0.75}
\definecolor{brinkpink}{rgb}{0.98, 0.38, 0.5}
\definecolor{aogreen}{rgb}{0.0, 0.5, 0.0}
\definecolor{myorange}{rgb}{1.0, 0.49, 0.0}
\definecolor{cadetgrey}{rgb}{0.5, 0.5, 0.5}

\def\showcomments{1}
\if\showcomments1
\newcommand{\gn}[1]{\textcolor{magenta}{\bf\small [#1 --GN]}}

\newcommand{\pfliu}[1]{\textcolor{cyan}{\bf\small [#1 --pfliu]}}

\else 
\newcommand{\gn}[1]{}

\newcommand{\pfliu}[1]{}

\fi

\usepackage[font={small}]{caption}
\usepackage{subfig}
\usepackage{graphicx} 
\usepackage{booktabs}       
\usepackage{amsfonts}       
\usepackage{nicefrac} 
\usepackage{mathrsfs}
\usepackage{amsfonts}
\usepackage{amssymb}
\usepackage{amsmath}
\usepackage{bm}
\usepackage{url}
\usepackage{booktabs}
\usepackage{times}
\usepackage{soul}
\usepackage{hyperref}
\usepackage{url}
\usepackage{helvet}
\usepackage{courier}
\usepackage{graphicx} 
\usepackage{arydshln}
\usepackage{multirow}
\usepackage{multicol}
\usepackage{hyperref}
\usepackage{booktabs}
\usepackage{booktabs}       
\usepackage{amsfonts}       
\usepackage{nicefrac}       
\usepackage{microtype}      
\usepackage{verbatim}
\usepackage{amssymb}
\usepackage{pifont}

\usepackage{wrapfig}
\usepackage{lipsum}

\aclfinalcopy 

\newcommand*\samethanks[1][\value{footnote}]{\footnotemark[#1]}

\newenvironment{itemize*}%
  {\begin{itemize}%
    \setlength{\itemsep}{0pt}%
    \setlength{\parskip}{0pt}}%
  {\end{itemize}}
  \newenvironment{enumerate*}%
  {\begin{enumerate}%
    \setlength{\itemsep}{0pt}%
    \setlength{\parskip}{0pt}}%
  {\end{enumerate}}

\title{Interpretable Multi-dataset Evaluation for \\ Named Entity Recognition}

\author{Jinlan Fu\\
  Affiliation / Address line 1 \\
  Affiliation / Address line 2 \\
  Affiliation / Address line 3 \\
  \texttt{email@domain} \\\And
  Pengfei Liu \\
  Affiliation / Address line 1 \\
  Affiliation / Address line 2 \\
  Affiliation / Address line 3 \\
  \texttt{email@domain} \\}

\date{}

\author{Jinlan Fu$\dag$ \thanks{\ \  These two authors contributed equally.}, \quad Pengfei Liu$\sharp$ \samethanks, \quad Graham Neubig$\sharp$ \\
   $\dag$ Fudan University, $\sharp$Carnegie Mellon University \\
  \texttt{fujl16@fudan.edu.cn}, \texttt{\{pliu3,gneubig\}@cs.cmu.edu}}

\begin{document}
\maketitle

\begin{abstract}

    With the proliferation of models for natural language processing tasks, it is even harder to understand the differences between models and their relative merits. Simply looking at differences between \emph{holistic} metrics such as accuracy, BLEU, or F1 does not tell us \emph{why} or \emph{how} particular methods perform differently and how diverse datasets influence the model design choices.
    In this paper, we present a general methodology for \emph{interpretable} evaluation for the named entity recognition (NER) task.
    The proposed evaluation method enables us to interpret the differences in models and datasets, as well as the interplay between them, identifying the strengths and weaknesses of current systems.
    By making our analysis tool available, we make it easy for future researchers to run similar analyses and drive progress in this area: \url{https://github.com/neulab/InterpretEval}. 
\end{abstract}

\section{Introduction}
With improvements in model architectures \cite{hochreiter1997long,kalchbrenner2014convolutional,lample2016neural,collobert2011natural} and learning of pre-trained embeddings \cite{peters2018deep,akbik2018contextual,akbik2019pooled,devlin2018bert,pennington2014glove}, Named Entity Recognition (NER) systems are evolving rapidly but also quickly reaching a performance plateau \citep{akbik2018contextual,akbik2019pooled}.
This proliferation of methods poses a great challenge for the current evaluation methodology, which usually is based on comparing systems on a single holistic score assessing accuracy (usually entity $F1$-score).
There are several issues with this practice.
First, a single evaluation number does not allow us to distinguish on a fine-grained level the strengths and weaknesses among diverse systems.
Second, it is hard to improve what we do not understand; if an engineer or researcher looking to make model improvements cannot tell where the model is failing, it is also hard to decide which methodological improvements to try next.

To alleviate this problem,
a few works \citep{ichihara2015error,derczynski2015analysis} have attempted to perform \emph{fine-grained error analysis} of NER systems.
While a step in the right direction, these analyses frequently 
rely upon labor-intensive manual examination and also customarily depend on pre-existing error typologies encoding assumptions about the errors a system is likely to make.

\begin{figure}%
\centering
\includegraphics[width=0.98\linewidth]{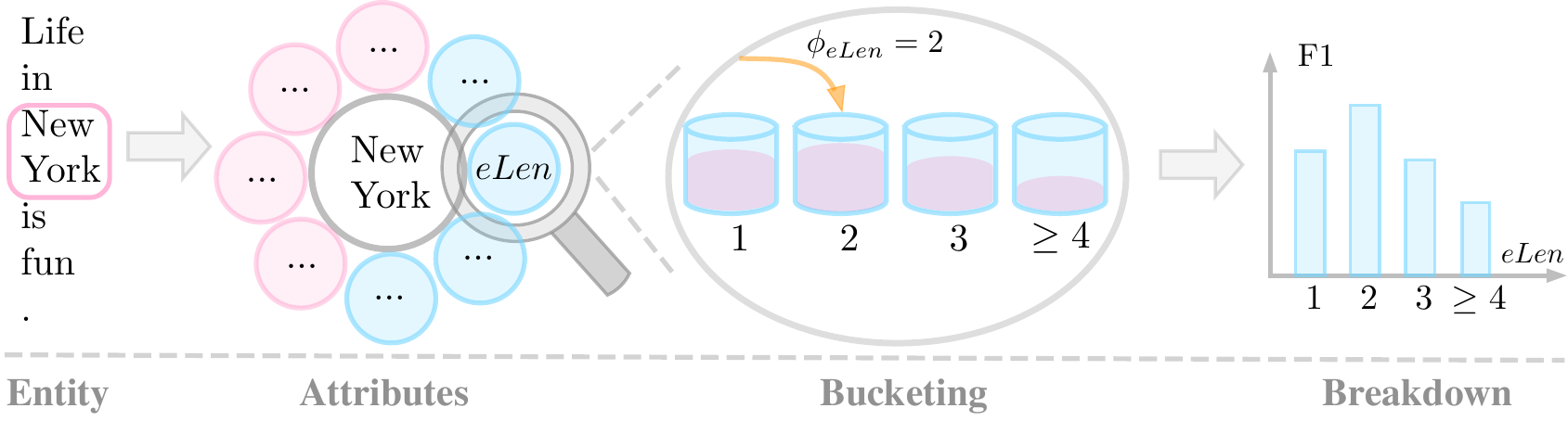}
\vspace{-6pt}
\caption{An example of our evaluation methodology. \texttt{eLen} (\textit{entity length}) represents one of the attributes (detailed in Sec.~\ref{sec:attribute-definition}) of the entity ``\texttt{New York}''.
After bucketing, performance can be broken down over different attribute values.
} 
\label{fig:new_york}
\end{figure}

Orthogonally, some other works~\cite{qian2018graphie:,hu2020leveraging,luo2020hierarchical,li2019a,TriggerNER2020} evaluate holistic metrics such as F1 \textit{across multiple datasets} that differ in domain, language, or other characteristics \citep{sang2003introduction,collobert2011natural,weischedel2013ontonotes}.
Although this enables us to more comprehensively assess the models, the reliance on holistic metrics precludes a finer-grained view of how various aspects of the model performance vary across the different settings.

In this paper, we argue that an ideal evaluation methodology should be (1) fully or partially automatic, (2) allow evaluation and comparison across multiple datasets, and (3) allow users to dig deeper into fine-grained strengths and weaknesses of each model. 
To this end, we devise a generalized, fine-grained, and multi-dataset evaluation methodology for the task of NER, as demonstrated in Fig.~\ref{fig:new_york}.
Specifically, 
it leverages the notion of ``\textit{attributes}'', values which characterize the properties of an entity that may be correlated with the NER performance (e.g.~entity length in words).
Afterward, we partition test entities into a set of \textit{buckets} based on the entity's attributes, where entities in different buckets may have different performance scores on average (e.g.~entities with more words may be predicted less accurately).

\begin{figure}%
\centering
\includegraphics[width=0.55\linewidth]{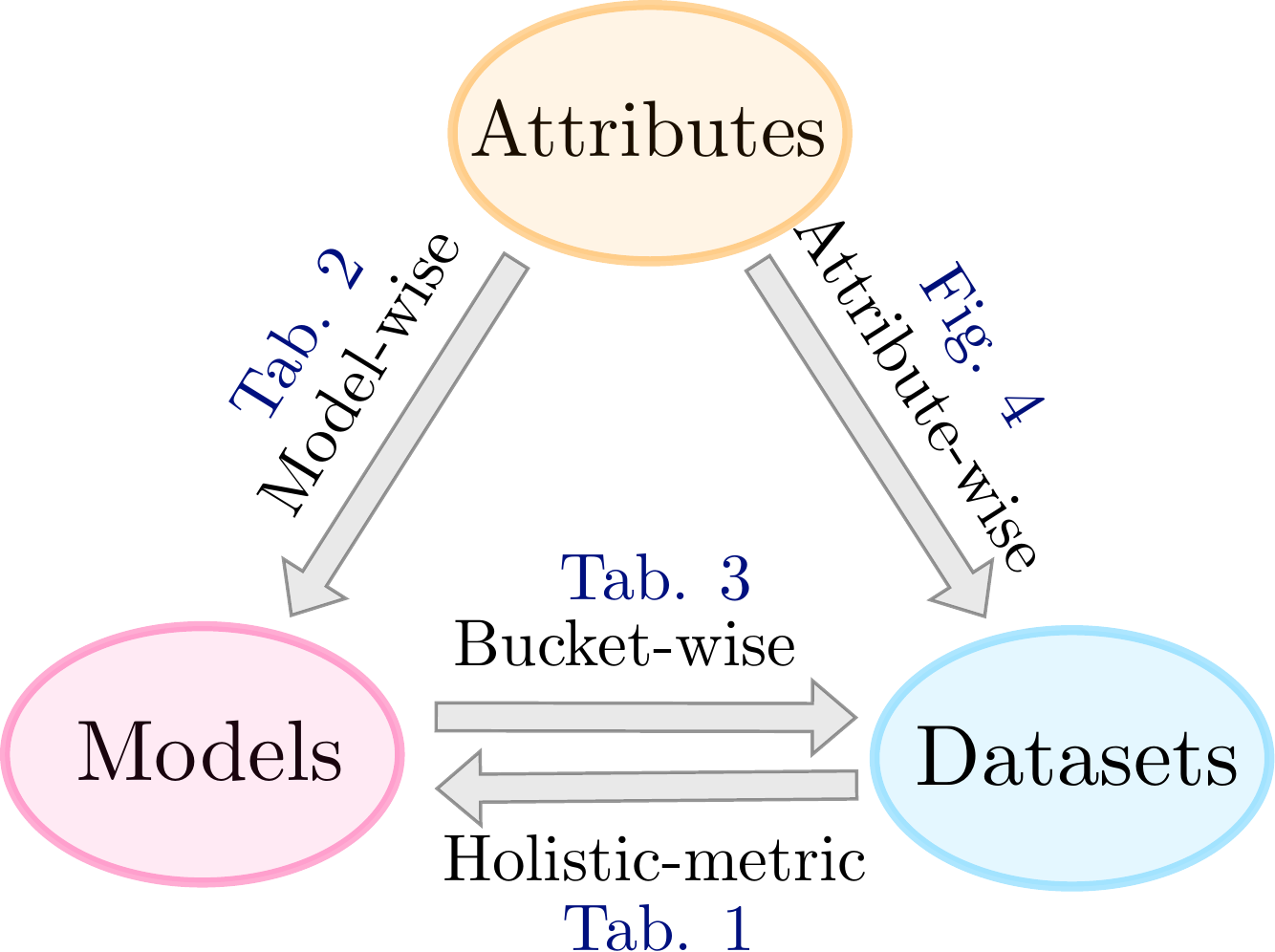}
\vspace{-6pt}
\caption{Relations among attributes, models, and datasets.}
\label{fig:MAD_relation}
\end{figure}

Methodologically, our evaluation framework allows for three analytical views as elucidated in Fig.~\ref{fig:MAD_relation}.
\textit{Model-wise} (Sec.~\ref{sec:model-wise}) analysis investigates how the performances of different models vary according to attribute value (e.g.~``Is a model with a CRF layer better at dealing with long entities?'').
\textit{Attribute-wise}
(Sec.~\ref{sec:attribute-wise}) analysis compares how different attributes affect performance on different datasets (e.g.~``Does entity length correlate with model performance on all datasets or just some?''). 
\textit{Bucket-wise} (Sec.~\ref{sec:bucket-wise}) compares among all possible analysis dimensions, and can diagnose the strengths and weaknesses of existing models (e.g.~``What entity attributes indicate that a BERT-based model will likely fail?''),
or help us understand how different choices of datasets influence model performance (e.g.~``On which datasets is using a CRF layer more appropriate?'').

Experimentally, we conduct a comprehensive analysis over \textbf{twelve} models, \textbf{eight} attributes, and \textbf{six} datasets. Proposed quantifiable measures allow us to draw several qualitative conclusions as highlighted below:
1) \textit{label consistency} (the degree of label agreement of an entity on the training set) and \textit{entity length} have a consistent influence on NER task's performance (Sec.~\ref{sec:attribute-wise-sub}); 
2) CRF-based systems are more likely to make a mistake compared with MLP-based systems when dealing with long entities (Sec.~\ref{sec:model-wise-analysis});
3) Higher-frequency tokens of a test entity cannot guarantee better performance since other crucial factors such as \textit{label consistency} also matter (Sec.~\ref{sec:attribute-wise-sub});
4) Even more advanced models (e.g., \texttt{BERT}, \texttt{Flair}) fail to predict entities with low \textit{label consistency} (Sec.~\ref{sec:self-diagnosis-analysis}).

Finally, motivated by observation 4), we present an effective solution to improve current NER systems.
Quantitative and qualitative experiments demonstrate that introducing larger context is an effective method, obtaining improvements of up to 10 points in F1 score on some datasets.

\section{Background}

\paragraph{Task Description} 
NER is frequently formulated as a sequence labeling problem \citep{chiu2015named,huang2015bidirectional,ma2016end,lample2016neural}, where $X = \{x_1,x_2,\ldots, x_T\}$ is an input sequence and $Y = \{y_1,y_2,\ldots,y_T\}$ are the output labels (e.g., ``B-PER'', ``I-LOC'', ``O'').
The goal of this task is to accurately predict entities by assigning output label $y_t$ for each token $x_t$:
$P(Y | X) = P(y_t|X,y_1,\cdots,y_{t-1})$

\paragraph{Standard Evaluation Strategy for NER}
The common evaluation metric for NER systems \cite{sang2003introduction} is to compute a corpus-level metric using micro-averaged $F1$ score: $F1 = \frac{2\times P\times R}{P+R}$: 
where $P$ is the percentage of named entities output by learning system that are correct. $R$ is the percentage of gold entities identified by the system.
Here a named entity is correct only if it is an exact match of an annotated entity.

\section{Attribute-aided Evaluation}
Our proposed attribute-aided evaluation methodology involves two key elements: 
\textit{attribute definition} and \textit{bucketing}.
We first define diverse attributes for each entity, by which test entities are partitioned into different buckets.
We then calculate the performance for each bucket of test entities.

\subsection{Attribute Definition} \label{sec:attribute-definition}
Attributes are defined either over a \emph{span} or a \emph{token} and characterize the diverse properties thereof.
In practice, the span will be instantiated as a genuine or a mis-predicted entity (calculating \textit{precision}) in the test set, while tokens can be any token in the test corpus.
We classify attributes into two categories: local attributes and aggregate attributes.

\paragraph{Local attributes} are calculated with respect to a span or token regarding attributes of the span/token itself, its label, or the sentence in which the span appeared.
We define a token $x$ or span $\mathbf{x}$, to have a gold-standard or predicted label $y=\text{lab}(\cdot)$,\footnote{$y$ is a simple entity label for tokens, and does not distinguish between ``B'' and ``I'' in the BIO tagging scheme.} which occurs in sentence $X=\text{sent}(\mathbf{x})$.
We also define two functions that count the number of words outside the training set\footnote{Not considering the vocabulary of pre-trained models.} 
$\text{oov}(\cdot)$ and the number of entities $\text{ent}(\cdot)$ in a sequence of words. 
Based on this, we define several feature functions $\phi(\mathbf{x})$ that can compute different attributes of each span:
\vspace{-5pt}
\begin{itemize*}
    \item  $\phi_{\texttt{str}}(\mathbf{x})$ = $\mathbf{x}$: \emph{span surface string}
    \item  $\phi_{\texttt{label}}(\mathbf{x}) = \text{lab}(\mathbf{x})$: \emph{entity span label}      
    \item  $\phi_{\texttt{eLen}}(\mathbf{x}) = |\mathbf{x}|$: \emph{entity span length}    
\end{itemize*}
\vspace{-5pt}
We additionally define feature functions over tokens $\phi(x)$:
\vspace{-5pt}
\begin{itemize*}
    \item  $\phi_{\texttt{str}}(x)$ = $x$: \emph{token surface string}
    \item  $\phi_{\texttt{label}}(x) = \text{lab}(x)$: \emph{token label}     
\end{itemize*}
\vspace{-5pt}
We also define several features of the underlying sentence, which can be applied to either spans $\mathbf{x}$ or tokens $x$; we show the example of applying to token $x$ below:
\vspace{-5pt}
\begin{itemize*}
    \item $\phi_{\texttt{sLen}}(x) = |\text{sent}(x)|$: \emph{sentence length}
    \item $\phi_{\texttt{eDen}}(x) = |\text{ent}(\text{sent}(x))|/\phi_{sLen}(x)$: \\ \emph{entity density}   
    \item $\phi_{\texttt{oDen}}(x) = |\text{oov}(\text{sent}(x))|/\phi_{sLen}(x)$: \\ \emph{OOV density}  
\end{itemize*}
\vspace{-5pt}

\begin{figure}
\centering
\includegraphics[width=0.98\linewidth]{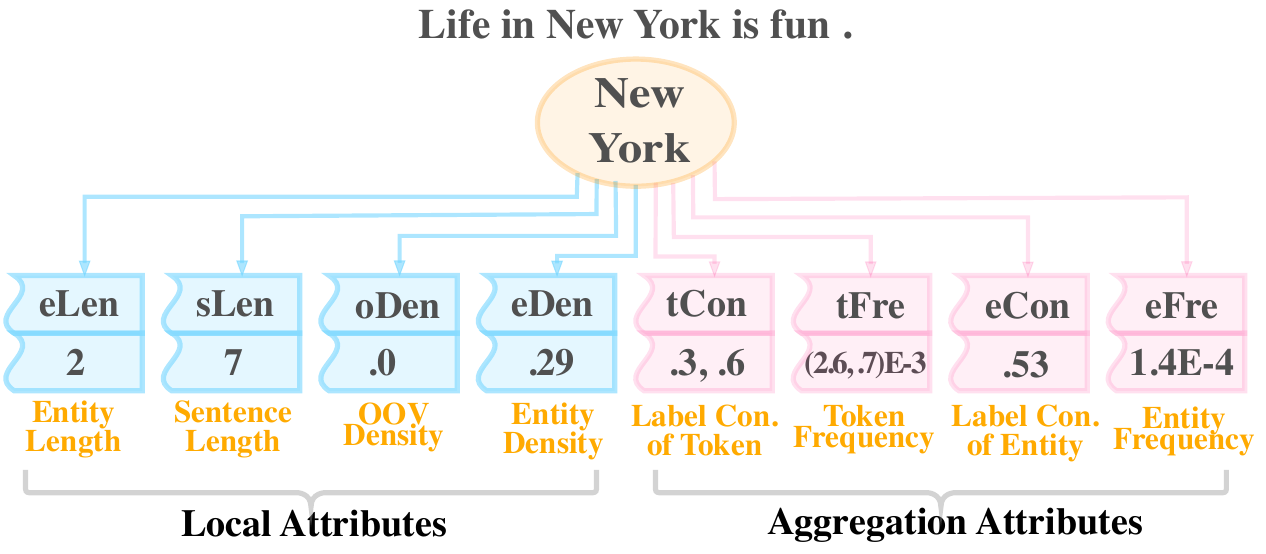}
\vspace{-6pt}
\caption{The eight attributes defined in this paper and corresponding values with respect to the entity ``\texttt{New York}'' in the sentence ``\texttt{Life in New York is fun .}''.  The text in orange is the full name of the attribute, in which \textit{Con.} denotes \textit{Consistency}.
} 
\label{fig:attributes}
\end{figure}

\paragraph{Aggregate attributes} are properties of spans or tokens based on aggregate statistics that require calculation over the whole training corpus.
To calculate these attributes, we first define $\mathcal{E}^{\text{tr}}$ as all spans/tokens in the training set.
We then define an aggregation function that takes a particular span/token (example of tokens below), feature function $\phi(\cdot)$, and span set $\mathcal{E} \subseteq \mathcal{E}^{\text{tr}}$ as arguments:

\vspace{-10pt}
{\small
\begin{align}
    \mathcal{F}(x, \phi(\cdot), \mathcal{E}) = \frac{|\{\varepsilon| \phi(\varepsilon) = \phi(x) , \forall \varepsilon \in \mathcal{E}\}|}{|\mathcal{E}|}, \label{eq:fbucket}
\end{align}
}
\vspace{-18pt}

\noindent calculating the ratio of spans/tokens in $\mathcal{E}$ that have the same feature value $\phi(\cdot)$ as $x$.
We can define $\mathcal{E} := \mathcal{E}^{\text{tr}}$, calculating statistics over the entire training set.
We can also choose it to be only the spans/tokens with a particular surface form:

\vspace{-15pt}
\begin{equation}
    \mathcal{E}^{x} := \{\varepsilon|\phi_{\texttt{str}}(\varepsilon)=\phi_{\texttt{str}}(x), \varepsilon\in \mathcal{E}^{\text{tr}}\}.
\end{equation}
\vspace{-20pt}

Based on the above general formulation, we defined a few specific instantiations that we use in the following experiments.
First, \emph{entity frequency} and \emph{token frequency}:

\vspace{-20pt}
\begin{align}
    \phi_{\texttt{eFre}}(\mathbf{x}) & := \mathcal{F}(\mathbf{x}, \phi_{\texttt{str}}(\cdot), \mathcal{E}^{\text{tr}})  \label{eq:efre} \\
    \phi_{\texttt{tFre}}(x) & := \mathcal{F}(x, \phi_{\texttt{str}}(\cdot), \mathcal{E}^{\text{tr}})  \label{eq:tfre}
\end{align}
\vspace{-20pt}

Besides, we use two \emph{consistency}-based attributes, which attempt to measure how consistently a particular span/token is labeled with a particular label:

\vspace{-20pt}
\begin{align}
    \phi_{\texttt{eCon}}(\mathbf{x}) & := \mathcal{F}(\mathbf{x}, \phi_{\texttt{label}}(\cdot), \mathcal{E}^{\text{tr}})  \label{eq:econ} \\
    \phi_{\texttt{tCon}}(x) & := \mathcal{F}(x, \phi_{\texttt{label}}(\cdot), \mathcal{E}^{\text{tr}})  \label{eq:tcon}
\end{align}
\vspace{-20pt}

We give an example to illustrate above  by setting $\mathbf{x} = $ ``\texttt{New York}'' with gold label ``\texttt{LOC}''. Therefore, the numerator of $\phi_{eCon}$ tallies entities ``\texttt{New York}'' with label ``\texttt{LOC}'' in training set, while the denominator counts spans ``\texttt{New York}''. The overall ratio quantifies the degree of label consistency in train set for a given span ``\texttt{New York}''.

\subsection{Bucketing} \label{sec:bucket-strategy}
\textit{Bucketing} is an operation that breaks down the holistic performance into different categories~\cite{neubig2019compare,fu2020rethinking}.
This can be achieved by dividing the set of test entities into different subsets of test entities (regarding span- and sentence-level attributes) or test tokens (regarding token-level attributes). 
Here we describe the entity-based bucketing strategies, which can also be similarly applied to token-based strategies. 
The bucketing process can be expressed in the following general form: 

\vspace{-12pt}
\begin{equation}
    \mathcal{E}_1^{te}, \cdots, \mathcal{E}_m^{te} = \mathrm{Bucket}(\mathcal{E}^{te}, \phi(\cdot)) 
\end{equation}
\vspace{-18pt}

where $\mathcal{E}^{te}$ represents a set of test entities or tokens, and $m$ is the number of buckets. $\phi(\cdot)$ denotes one type of feature functions (as defined in Sec.~\ref{sec:attribute-definition}) to calculate attribute value for a given entity (e.g., $\phi_{\texttt{eLen}}(\mathbf{x})$ to compute \textit{span length}).

Specifically, we divide the range of attribute values into $m$ discrete parts, whose intervals can be obtained mainly based on two ways: 1) dividing value range evenly  2) dividing test entities or tokens equally.
In practice, the way to obtain intervals may be diverse for different attributes.%
\footnote{We have implemented flexible functions to do this as users need in our released code.}
We detail our settings in the appendix.
Finally, once we have generated buckets, we calculate the F1 score with respect to entities (or tokens) of each bucket.  
We can easily extend the attribute-aided evaluation to other tasks, such as Chinese Word Segmentation~\cite{fu2020rethinkcws}.

\section{Experimental Settings}

\begin{table*}[htb]
\renewcommand\tabcolsep{2pt}
  \centering \scriptsize
    \begin{tabular}{lcccccccccccccccccc}
    \toprule
    \multicolumn{1}{c}{\multirow{2}[4]{*}{\textbf{Models}}}
    & \multicolumn{5}{c}{\textbf{Char/Subword }} & \multicolumn{3}{c}{\textbf{Word}} & \multicolumn{2}{c}{\textbf{Sentence}} & \multicolumn{2}{c}{\textbf{Decoder}} &
    \multicolumn{6}{c}{\textbf{Overall F1}}\\
\cmidrule(lr){2-6}\cmidrule(lr){7-9}\cmidrule(lr){10-11}\cmidrule(lr){12-13} \cmidrule(lr){14-19}
& \textbf{\rotatebox{90}{none}} & \textbf{\rotatebox{90}{cnn}} & \textbf{\rotatebox{90}{elmo}} & \textbf{\rotatebox{90}{flair}} & \textbf{\rotatebox{90}{bert}} & \textbf{\rotatebox{90}{none}} & \textbf{\rotatebox{90}{rand}} & \textbf{\rotatebox{90}{glove}} & \textbf{\rotatebox{90}{lstm}} & \textbf{\rotatebox{90}{cnn}} & \textbf{\rotatebox{90}{crf}} & \textbf{\rotatebox{90}{mlp}} &
\multicolumn{1}{c}{\textbf{CoNLL}} & \multicolumn{1}{c}{\textbf{WNUT}} & \multicolumn{1}{c}{\textbf{BN}} & \multicolumn{1}{c}{\textbf{BC}} & \multicolumn{1}{c}{\textbf{MZ}} & \multicolumn{1}{c}{\textbf{WB}} 
\\
    \midrule
    \textit{CRF++} 
    &&&&&&&&&&&&& 80.74 &21.53 &82.02 &67.71 &77.80 &47.90\\
    \textit{CnonWrandLstmCrf} & $\surd$     &       &       &       &       &       & $\surd$     &       & $\surd$    &       & $\surd$     & &  78.13 & 17.24 & 80.36 & 66.17 & 73.89 & 49.80 \\
    \textit{CcnnWnoneLstmCrf} &       & $\surd$     &       &       &       & $\surd$     &       &       & $\surd$     &       & $\surd$     & & 77.01 & 22.73 & 77.96 & 65.01 & 79.05 & 47.31 \\
    \textit{CcnnWrandLstmCrf} &       & $\surd$     &       &       &       &       & $\surd$      &       & $\surd$      &       & $\surd$      &  & 83.80  & 22.57 & 83.59 & 71.57 & 78.85 & 52.14 \\

    \midrule
    \textit{CcnnWgloveLstmCrf} &       & $\surd$      &       &       &       &       &     & $\surd$      &        $\surd$   &   & $\surd$      & &  90.48 & 40.61 & 86.78 & 76.04 & 85.39 & 60.17 \\
    \textit{CcnnWgloveCnnCrf} &       & $\surd$      &       &       &       &    &   & $\surd$ &  & $\surd$ &  $\surd$ & & 90.14 & 36.21 & 86.42 & 76.74 &  \textbf{88.10}  & 49.10 \\
    \textit{CcnnWgloveLstmMlp} &       & $\surd$      &       &       &       &       &       & $\surd$      & $\surd$      &   &    & $\surd$       & 88.05 & 32.84 & 84.07 & 70.00    & 81.09 & 56.61 \\
    \midrule
    \textit{CelmWnoneLstmCrf} &       &       & $\surd$      &       &       & $\surd$      &       &       & $\surd$      &       & $\surd$    &  &  91.64 & 44.56 & \textbf{89.75} & 77.10  & 86.32 & 60.51 \\
    \textit{CelmWgloveLstmCrf} &       &       & $\surd$      &       &       &       &       & $\surd$      & $\surd$      &       & $\surd$     &  & 92.22 & 45.33 & 89.35 & 78.71 & 85.70  & 63.26 \\
    \textit{CbertWnoneLstmMlp} &       &       &       &       & $\surd$      & $\surd$      &       &       & $\surd$      &       &   & $\surd$   &  91.11 & \textbf{47.77}  & 89.64 & \textbf{81.03} & 86.90  & \textbf{66.35} \\
    \textit{CflairWnoneLstmCrf} &       &       &       & $\surd$      &       & $\surd$      &       &       & $\surd$      &       & $\surd$ &     &  89.98 & 41.49 & 87.98 & 77.46 & 84.11 & 56.71 \\
    \textit{CflairWgloveLstmCrf} &       &       &       & $\surd$      &       &       &       & $\surd$      & $\surd$      &       & $\surd$ &     &  \textbf{93.03} & 45.96 & 87.92 & 77.23 & 85.56 & 63.38 \\
    
    \bottomrule
    \end{tabular}%
    \vspace{-5pt}
   \caption{Neural NER systems with different architectures. CRF++ is a Conditional Random Fields~\cite{lafferty2001conditional} method based on feature engineering.
   Bold is the best performance of a given dataset according to F1. 
   For the model name, ``\texttt{C}" refers to ``\texttt{Char/Subword}" and ``\texttt{W}" refers to ``\texttt{Word}".
   For example, "\textit{CnonWrandLstmCrf}" is a model with no character features, with randomly initialized embeddings, and the sentence encoder is LSTM and decoder is CRF. 
   } 
    \label{tab:allmodels}%
\end{table*}%

In this section we describe our experimental settings, each of which is followed by an experiment and analysis results.

\subsection{NER Datasets for Evaluation}

We conduct experiments on: CoNLL-2003 (\texttt{CoNLL}),~\footnote{https://www.clips.uantwerpen.be/conll2003/ner/}  WNUT-2016 (\texttt{WNUT}),\footnote{https://noisy-text.github.io/2016/ner-shared-task.html}  and OntoNotes 5.0 dataset.~\footnote{https://catalog.ldc.upenn.edu/LDC2013T19} 
The CoNLL dataset \citep{sang2003introduction} is based on Reuters data \citep{collobert2011natural}. The WNUT dataset~\citep{strauss2016results} is provided by the second shared task at WNUT-2016 and consists of social media data from Twitter. The OntoNotes 5.0 dataset  \citep{weischedel2013ontonotes} is collected from broadcast news (\texttt{BN}), broadcast conversation (\texttt{BC}), weblogs (\texttt{WB}), and magazine genre (\texttt{MZ}).

\subsection{Models}
We varied the evaluated models mainly in terms of four aspects:
1) character/subword-sensitive encoder: \texttt{ELMo} \citep{peters2018deep}, \texttt{Flair} \citep{akbik2018contextual,akbik2019pooled}, \texttt{BERT}~\footnote{The reason why we group BERT into a subword-sensitive encoder is that we use it to obtain the representation of each subword.} \citep{devlin2018bert} 

2) additional word embeddings: \texttt{GloVe} \citep{pennington2014glove};
3) sentence-level encoders: \texttt{LSTM} \citep{hochreiter1997long}, \texttt{CNN} \citep{kalchbrenner2014convolutional,chen2019grn:};
4) decoders: \texttt{MLP} or \texttt{CRF} \citep{lample2016neural,collobert2011natural}. 
In total, we study 12 NER models and we give more detailed description of models in the appendix.  
Detailed model settings are illustrated in Tab.\ref{tab:allmodels}.
We use the result from the model with the best development set performance, terminating training  when the performance on development is not improved in 20 epochs.

\subsection{Holistic Analysis}
\label{sec:holistic-analysis}
Before giving a fine-grained analysis, we present the results of different models on different datasets as traditional \textit{multi-dataset} evaluation does.
As Tab.~\ref{tab:allmodels} demonstrates, 
there is no one-size-fits-all model; different models get the best results on different datasets.
Naturally, this raises the following questions:
1) what factors of the datasets significantly influence NER performance?
2) how do these factors influence the choices of models?
3) does a worse-ranked model outperform the best-ranked model in some aspects and how do datasets influence the choices of models?
The following analyses will investigate these questions.

\section{Fine-grained Analysis}
To better characterize the relationship among models, attributes, and datasets, we propose three analysis approaches: model-, attribute-, and bucket-wise.

Formally, we refer to $M = \{m_1,\cdots, m_{|M|}\}$ as a set of \textbf{models} and
$ \Phi  =\{\phi_1,\cdots,\phi_{|\Phi|} \}$ 
as a set of \textbf{attributes}.
As described in Sec.~\ref{sec:bucket-strategy}, the test set $\mathcal{E}$ could be split into different \textbf{buckets} of $\mathcal{E} =\{ \mathcal{E}^j_{1}, \cdots, \mathcal{E}^j_{|\mathcal{E}|} \}$ based on an attribute $\phi_j$.
We introduce the concept of a \textit{performance table} $\mathcal{T} \in \mathbb{R}^{|M|\times |\Phi| \times |\mathcal{E}|}$, whose element $\mathcal{T}_{ijk}$ represents the performance ($F1$ score) of $i$-th model on the $k$-th sub-test set (bucket) generated by $j$-th attribute. 
Next, we will explain how above-mentioned analysis approaches are defined based on $\mathcal{T}$.

\renewcommand\tabcolsep{0.6pt}

\begin{table}[!ht]
  \centering  \tiny 
    \begin{tabular}{lccccccccccccccccccc}
    \toprule 
          & & \multicolumn{8}{c}{\textbf{Spearman ($\mathbf{S}^{\rho}_{i,j}$)}}                                 &  \multicolumn{10}{c}{\textbf{Standard Deviation ($\mathbf{S}^{\sigma}_{i,j}$)} }\\
    \cmidrule(lr){3-10}\cmidrule(lr){11-20}
    \textbf{Model} & \textbf{F1} & \rotatebox{90}{eDen} & \rotatebox{90}{oDen} & \rotatebox{90}{sLen} & \rotatebox{90}{eCon} & \rotatebox{90}{tCon} & \rotatebox{90}{eFre} & \rotatebox{90}{tFre} & \rotatebox{90}{eLen} & \rotatebox{90}{eDen} & \rotatebox{90}{oDen} & \rotatebox{90}{sLen} & \rotatebox{90}{eCon} & \rotatebox{90}{tCon} & \rotatebox{90}{eFre} & \rotatebox{90}{tFre} & \rotatebox{90}{eLen}   \\
    \midrule
    \textit{CRF++} & \textcolor[rgb]{0.0, 0.5, 0.0}{55.00}  & \textcolor{cadetgrey}{-4}  & \textcolor{cadetgrey}{9}     & \textcolor{cadetgrey}{-10}          & \textcolor[rgb]{0.0, 0.5, 0.0}{87}    & \textcolor[rgb]{0.0, 0.5, 0.0}{79}    & 96    & 56  & \textcolor[rgb]{0.98, 0.38, 0.5}{\textbf{-92 }} & \textcolor{cadetgrey}{5.5}   & \textcolor{cadetgrey}{7.5}   & \textcolor{cadetgrey}{5.2}      & \textcolor[rgb]{0.0, 0.5, 0.0}{16.2}  & \textcolor[rgb]{0.0, 0.5, 0.0}{12.7}  & 14.8  & 6.6   & \textcolor[rgb]{0.98, 0.38, 0.5}{5.8}  \\
    \textit{CnonWrandLstmCrf} & \textcolor[rgb]{0.0, 0.5, 0.0}{60.93} & \textcolor{cadetgrey}{-37}  & \textcolor{cadetgrey}{-2}    & \textcolor{cadetgrey}{-7}       & \textcolor[rgb]{0.0, 0.5, 0.0}{90}  & \textcolor[rgb]{0.0, 0.5, 0.0}{79}    & 94    & \textbf{57} & \textcolor[rgb]{0.98, 0.38, 0.5}{\textbf{-92 }} & \textcolor{cadetgrey}{5.9}   & \textcolor{cadetgrey}{8.2}   & \textcolor{cadetgrey}{4.4}      & \textcolor[rgb]{0.0, 0.5, 0.0}{\textbf{21.2}}  & \textcolor[rgb]{0.0, 0.5, 0.0}{16.3}  & 21.3  & 9.9   & \textcolor[rgb]{0.98, 0.38, 0.5}{7.8}  \\
    \textit{CcnnWnoneLstmCrf} & 61.51  & \textcolor{cadetgrey}{-11}  & \textcolor{cadetgrey}{-6}    & \textcolor{cadetgrey}{-7}     & 77    & \textbf{85 } & 95    & 49    & \textcolor[rgb]{0.98, 0.38, 0.5}{-75}   & \textcolor{cadetgrey}{6.1}   & \textcolor{cadetgrey}{6.7}   & \textcolor{cadetgrey}{5.6}       & 15.2  & 11.9  & 14.3  & 5.9   & \textcolor[rgb]{0.98, 0.38, 0.5}{7.2}  \\
    \textit{CcnnWrandLstmCrf} & \textcolor[rgb]{0.0, 0.5, 0.0}{65.42} & \textcolor{cadetgrey}{-19}   & \textcolor{cadetgrey}{5}     & \textcolor{cadetgrey}{-7}  & \textcolor[rgb]{0.0, 0.5, 0.0}{87}    & \textcolor[rgb]{0.0, 0.5, 0.0}{82}  & 95    & 44    & \textcolor[rgb]{0.98, 0.38, 0.5}{-92} & \textcolor{cadetgrey}{5.5}   & \textcolor{cadetgrey}{7.3}   & \textcolor{cadetgrey}{4.0}       & \textcolor[rgb]{0.0, 0.5, 0.0}{16.0}  & \textcolor[rgb]{0.0, 0.5, 0.0}{12.5}  & 15.5  & 6.6   & \textcolor[rgb]{0.98, 0.38, 0.5}{8.8}  \\
    \textit{CcnnWgloveLstmCrf} & 73.25 & \textcolor{cadetgrey}{-23}  & \textcolor{cadetgrey}{2}     & \textcolor{cadetgrey}{-15}        & 90  & 64    & 93    & 12    & \textcolor[rgb]{0.98, 0.38, 0.5}{\textbf{-92 }} & \textcolor{cadetgrey}{5.7}   & \textcolor{cadetgrey}{6.6}   & \textcolor{cadetgrey}{4.0}       & 12.0  & 9.2   & 14.9  & 5.2   & \textcolor[rgb]{0.98, 0.38, 0.5}{\textbf{9.0}}  \\
    \textit{CcnnWgloveCnnCrf} & 75.52 & \textcolor{cadetgrey}{-16}  & \textcolor{cadetgrey}{-11}   & \textcolor{cadetgrey}{-25}        & 90  & 65    & 88    & 0     & \textcolor[rgb]{0.98, 0.38, 0.5}{-83}   & \textcolor{cadetgrey}{5.6}   & \textcolor{cadetgrey}{6.8}   & \textcolor{cadetgrey}{3.8}       & 12.4  & 9.6   & 14.7  & 6.1  & \textcolor[rgb]{0.98, 0.38, 0.5}{\textbf{9.0}}  \\
    \textit{CcnnWgloveLstmMlp} & \textcolor[rgb]{0.98, 0.38, 0.5}{68.78} & \textcolor{cadetgrey}{-34}  & \textcolor{cadetgrey}{5}     & \textcolor{cadetgrey}{-17}         & \textbf{93 } & 63    & 97  & 3     & \textcolor[rgb]{0.98, 0.38, 0.5}{-67}   & \textcolor{cadetgrey}{5.9}   & \textcolor{cadetgrey}{6.8}   & \textcolor{cadetgrey}{3.9}       & 14.9  & 11.6  & 16.5  & 6.8   & \textcolor[rgb]{0.98, 0.38, 0.5}{7.1}  \\
    \textit{CelmWnoneLstmCrf} & 74.99 & \textcolor{cadetgrey}{7}  & \textcolor{cadetgrey}{3}     & \textcolor{cadetgrey}{5}          & 87    & 56    & \textbf{98 } & 16    & \textcolor[rgb]{0.98, 0.38, 0.5}{-83}   & \textcolor{cadetgrey}{5.8}   & \textcolor{cadetgrey}{6.6}   & \textcolor{cadetgrey}{4.1}       & 11.5  & 8.5   & 13.6  & 4.9   & \textcolor[rgb]{0.98, 0.38, 0.5}{6.5}  \\
    \textit{CelmWgloveLstmCrf} & 75.76 & \textcolor{cadetgrey}{-3}  & \textcolor{cadetgrey}{-8}    & \textcolor{cadetgrey}{-9}          & 87    & 60    & 93    & -2    &  \textcolor[rgb]{0.98, 0.38, 0.5}{\textbf{-92 }} & \textcolor{cadetgrey}{5.5}   & \textcolor{cadetgrey}{6.8}   & \textcolor{cadetgrey}{4.0}       & 11.4  & 8.2   & 13.4  & 5.1   & \textcolor[rgb]{0.98, 0.38, 0.5}{6.4}  \\
    \textit{CbertWnoneLstmMlp} & \textcolor[rgb]{0.98, 0.38, 0.5}{76.26} & \textcolor{cadetgrey}{0}  & \textcolor{cadetgrey}{-12}   & \textcolor{cadetgrey}{0}      & 83    & 56    & 87    & 17    & \textcolor[rgb]{0.98, 0.38, 0.5}{-58}   & \textcolor{cadetgrey}{5.4}   & \textcolor{cadetgrey}{5.6}   & \textcolor{cadetgrey}{3.7}       & 11.8  & 8.3   & 12.5  & 5.8   & \textcolor[rgb]{0.98, 0.38, 0.5}{4.8}  \\
    \textit{CflairWnoneLstmCrf} & 72.96 & \textcolor{cadetgrey}{-25}  & \textcolor{cadetgrey}{7}     & \textcolor{cadetgrey}{-23}         & 80    & 72    & 97  & 21    & \textcolor[rgb]{0.98, 0.38, 0.5}{-83}   & \textcolor{cadetgrey}{5.6}   & \textcolor{cadetgrey}{6.4}   & \textcolor{cadetgrey}{4.1}       & 12.2  & 9.1   & 13.6  & 5.3   & \textcolor[rgb]{0.98, 0.38, 0.5}{6.6}  \\
    \textit{CflairWgloveLstmCrf} & 75.51 & \textcolor{cadetgrey}{-16}  & \textcolor{cadetgrey}{-11}   & \textcolor{cadetgrey}{8}         & 87    & 67    & 91    & \textcolor{cadetgrey}{24}      & \textcolor[rgb]{0.98, 0.38, 0.5}{\textbf{-92 }} & \textcolor{cadetgrey}{5.2}   & \textcolor{cadetgrey}{5.8}   & \textcolor{cadetgrey}{4.0}       & 11.6  & 8.7   & 13.1  & \textcolor{cadetgrey}{5.3}      & \textcolor[rgb]{0.98, 0.38, 0.5}{6.5}  \\
    \bottomrule
    \end{tabular}%
    \vspace{-5pt}
    \caption{Model-wise measures (Percentage) $\mathbf{S}^{\rho}_{i,j}$ and $\mathbf{S}^{\sigma}_{i,j}$ which are the average over all the datasets. The F1 score for a model is also an average case on all the datasets.
    The value in grey denotes the attribute does not pass a significance test ($p\geq0.05$).
    The values in \textcolor[rgb]{0.0, 0.5, 0.0}{green} and in \textcolor[rgb]{0.98, 0.38, 0.5}{pink} support observation 1 and observation 2, respectively. The bold is the maximum value in the attribute column.
    }
  \label{tab:model-wise-conll}%
\end{table}%

\subsection{Exp-I: Model-wise Analysis}
\label{sec:model-wise}
Model-wise analysis investigates how different attributes influence performance of models with different architectures and initializations,  
e.g.~``does \texttt{eLen} influence performance of a \texttt{CNN-LSTM-CRF}-based NER system?''

\subsubsection{Approach}  \label{sec:model-wise-eq}
Here we adopt two types of statistical variables $\mathbf{S}^{\rho}_{i,j}$ and $\mathbf{S}^{\sigma}_{i,j}$ to characterize how the $j$-th attribute influences the of performance $i$-th model.
\vspace{-5pt}
\begin{align}
    \mathbf{S}^{\rho}_{i,j} &= \mathrm{Spearman}(\mathcal{T}[i,j:], R_{j}) \label{eq:m_rho} \\
    \mathbf{S}^{\sigma}_{i,j} &= \mathrm{Std}(\mathcal{T}[i,j:])  \label{eq:m_sigma}
\end{align}
where $\mathrm{Spearman}$  is a function to calculate the Spearman's rank correlation coefficient \citep{mukaka2012guide} and $R_{j}$ is the rank values of buckets based on the $j$-th attribute.  $\mathrm{Std}(\cdot)$ is the function to compute the standard deviation.

Intuitively, $\mathbf{S}^{\rho}_{i,j}$ characterizes how well the performance of the $i$-th model correlates with the values of the $j$-th attribute
while $\mathbf{S}^{\sigma}_{i,j}$ measures the degree to which this attribute influences the model's performance. 
For example, $\mathbf{S}^{\rho}_{CNN,\texttt{eCon}} = 0.9$ reveals that the performance of the \texttt{CNN} model positively and highly correlates with the attribute value \texttt{eCon} (label consistency). And a larger $\mathbf{S}^{\sigma}_{CNN,\texttt{eCon}}$ implies that \texttt{CNN} model's performance is heavily influenced by the factor \texttt{eCon}.

\noindent \textbf{Significance Tests}:
We perform Friedman's test \cite{zimmerman1993relative} with $p=0.05$. We examine whether the performance of different buckets partitioned by an attribute have the same expected performance, and the significance testing results are shown in appendix.
We omit the attributes whose $\mathbf{S}^{\rho}_{i,j}$ and  $\mathbf{S}^{\sigma}_{i,j}$ are not statistically significant (the values in grey in Tab.~\ref{tab:model-wise-conll}).

\subsubsection{Observations} 
\label{sec:model-wise-analysis}
Tab.~\ref{tab:model-wise-conll} illustrates the average case of $\mathbf{S}^{\rho}_{i,j}$ and $\mathbf{S}^{\sigma}_{i,j}$  on all datasets.\footnote{Correlations on individual datasets is in the Appendix.}
We highlight some major observations and more are in the appendix.

1) \textbf{The performance of character-unaware models is more sensitive to the label consistency.}
We observe that the performances of \textit{CRF++} and  \textit{CnonWrandLstmCrf} are highly correlated with \texttt{eCon}, and \texttt{tCon}   
with high values of $\mathbf{S}^{\rho}$ and $\mathbf{S}^{\sigma}$. 
Specifically, \textit{CcnnWrandLstmCrf} achieve higher performance and lower $\mathbf{S}^{\sigma}$ than \textit{CnonWrandLstmCrf}.
This suggests that the character-level encoder plays a major role in generalization to entities with low \textit{label consistency}.

2) \textbf{The influence of entity length varies greatly between different decoders.} 
Entity length is strongly negatively correlated with the performance of models, which means the performance of the model will drop with the entity length increasing. 
We observe that the variance scores $S^{\sigma}$ of \textit{CcnnWgloveLstmMlp} and \textit{CbertWnoneLstmMlp} are the smallest, compared with the variances of the models using non-contextualized and contextualized pre-trained embeddings, respectively.  
We attribute this to the structural biases of different decoders: \textit{MLP}-based models have better robustness when dealing with long entities, while \textit{CRF}-based models may lead to error propagation. We will present a detailed explanation of this in Sec.~\ref{sec:self-diagnosis-analysis}.

\subsection{Exp-II: Attribute-wise Analysis} \label{sec:attribute-wise}
Attribute-wise analysis aims to quantify the degree to which an attribute influences NER performance overall, across all systems.

\subsubsection{Approach}
\label{sec:attribute-wise-eq}

To achieve this, we introduce two dataset bias measures: task-independent variable $\zeta_j$ and task-dependent variable $\rho_j$ based on Eq.~\ref{eq:m_rho}:

\vspace{-15pt}
{\small
\begin{align}
    \zeta_j(\mathcal{E}, \phi(\cdot)) &= \frac{1}{N} \sum_i^{N} \phi_j(\mathbf{x}), 
    \label{eq:S_zeta} \\
    \rho_j &= \frac{1}{|M|} \sum_i^{|M|} |\mathbf{S}^{\rho}_{i,j}|, 
    \label{eq:S_rho1} 
\end{align}
}
\vspace{-15pt}

\noindent
where $\mathcal{E}$ denotes a dataset, $\phi_j(\mathbf{x})$ is the feature function to calculate an attribute value
for a given entity $\mathbf{x}$, $j$ denotes the $j$-th attribute function, and 
$N$ and $|M|$ are the numbers of test entities and models respectively.

For example, when $j$ denotes the attribute of sentence length, $\zeta_j$ is the average sentence length of the whole dataset. Intuitively, a higher absolute value of $\rho_j$ suggests that attribute $j$ is a crucial factor, whose values heavily correlate with the performances of NER systems. 

\begin{figure}[t]
    \centering
    
    \subfloat[$\zeta$]{
    \includegraphics[width=0.395\linewidth]{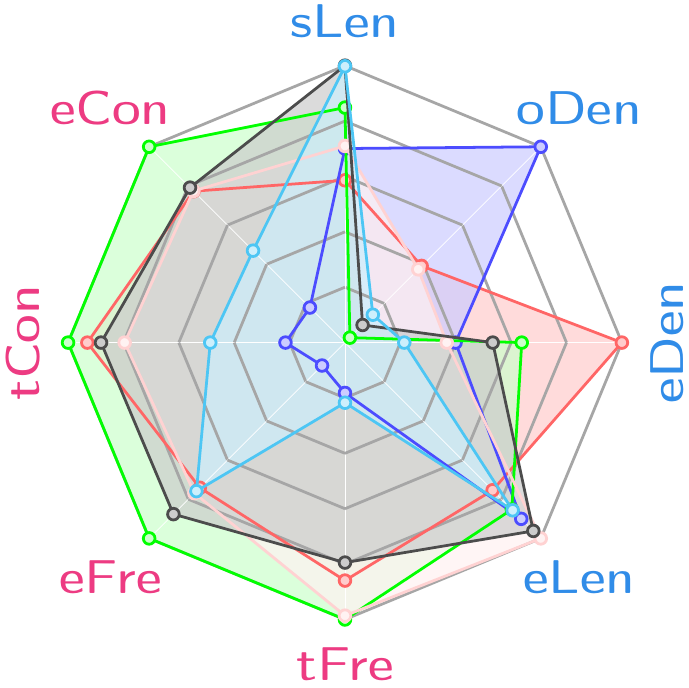} 
    }  \hspace{0.12em}  
    \subfloat[$\rho$ ]{
    \includegraphics[width=0.535\linewidth]{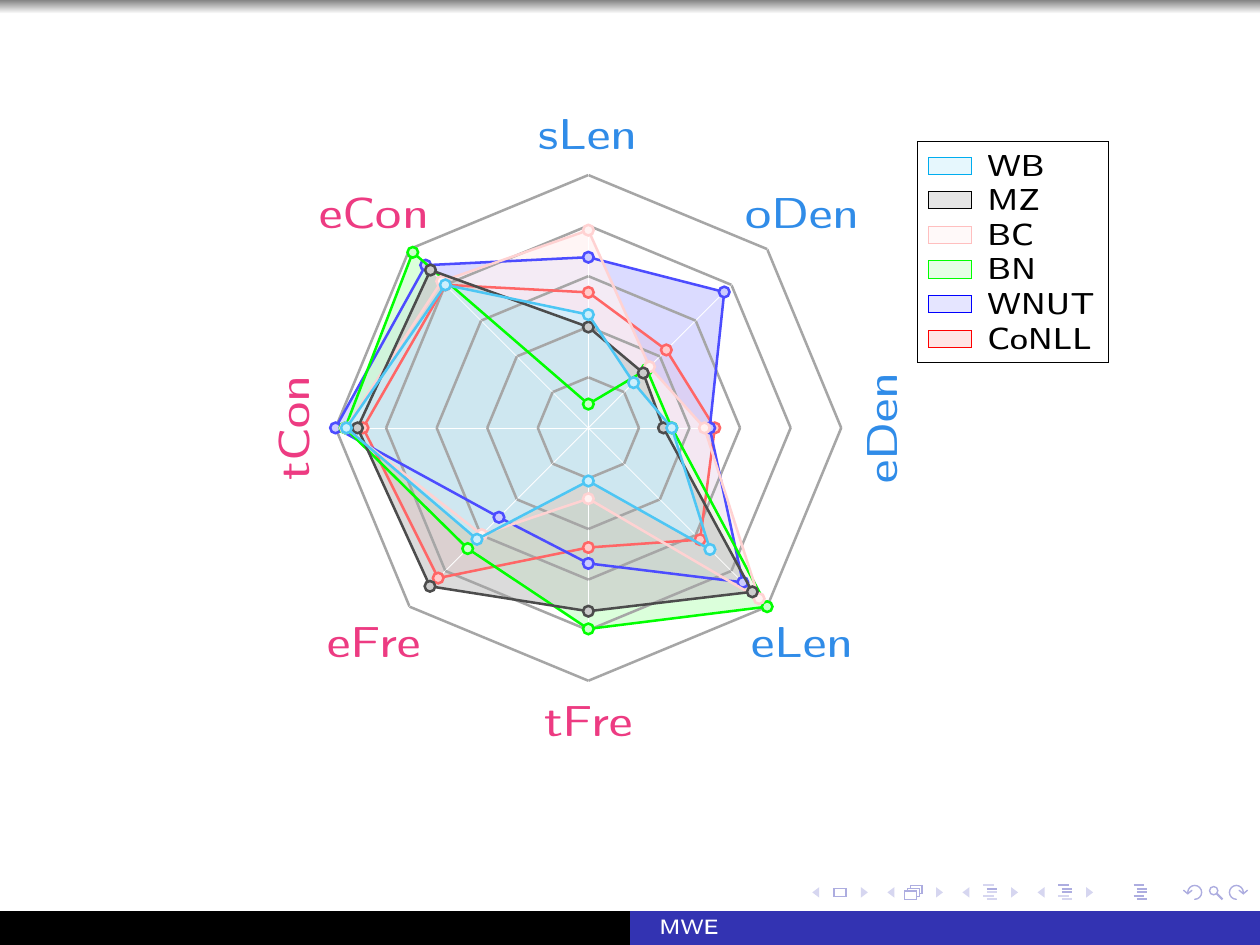} 
    }
    \vspace{-7pt}
    \caption{Dataset biases characterized by measures $\zeta$ and $\rho$. We normalize $\zeta$ on each attribute by dividing the maximum $\zeta$ on six datasets, and $\rho \in [0,1]$.}
    \label{fig:spider}
\end{figure}
\subsubsection{Observations}
\label{sec:attribute-wise-sub}
Similar to the above section, we conduct Friedman's test at $p= 0.05$. 
For all attributes, we find different-valued buckets are significantly different in their expected performance ($p < 0.05$). We include a full version of $p$ values in the appendix.
Detailed observations are listed as follows:

1) \textbf{Label consistency and entity length have a more consistent influence on NER performance.}
The common parts of the radar chart in Fig.~\ref{fig:spider}(b) illustrate that
for all datasets, the performance of the NER task is highly correlated with these attributes: \texttt{tCon} (label consisency of tokens), \texttt{eCon} (label consistency of entities), \texttt{eLen} (entity length). 
This reveals that the prediction difficulty of a named entity is commonly influenced by \textit{label consistency} (\texttt{tCon}, \texttt{eCon}) and \textit{entity length} (\texttt{eLen}).

2) \textbf{Frequency and sentence length matter but are minor factors.}
The outliers in the radar chart highlight the peculiarities of different datasets.
Intuitively, in Fig.~\ref{fig:spider}(b), on these attributes: \texttt{sLen}, \texttt{tFre}, \texttt{oDen}, the extent to which different datasets are affected varies greatly, and thus these attributes are not, in general, decisive factors for the NER task.  Typically, as observed from Fig.~\ref{fig:spider}(b), Spearman correlations of $\rho$ on the attribute \texttt{tFre} vary greatly, i.e., a smaller $\rho$ on \texttt{BC} and \texttt{WB}. This implies that \texttt{tFre} is not a decisive factor and \emph{higher-frequency tokens cannot guarantee better performance since other crucial factors such as \textit{label consistency} also matter}. We print the performance of the buckets with respect to token frequency, and find that the bucket with higher token frequency does not achieve a better performance.

Understanding these intrinsic differences in datasets provides us with evidence to explain how different datasets may influence different choices of models, which will be elaborated later (Sec.~\ref{sec:bucket-wise}).

\renewcommand\tabcolsep{0.9pt}
\renewcommand\arraystretch{0.73}  
\begin{table*}[!htb]
  \centering \tiny
    \begin{tabular}{ccccccc ccccccc ccccccc ccccccc ccccccc ccccccc ccccccc ccccccc ccccccc ccccccc ccccccc ccccccc ccccccc cccc} 
    \toprule
          & \multicolumn{15}{c}{CoNLL03}                           & \multicolumn{15}{c}{WNUT16}                            & \multicolumn{15}{c}{OntoNotes-MZ}                            & \multicolumn{15}{c}{OntoNotes-BC}  &
          \multicolumn{15}{c}{OntoNotes-BN}  &
          \multicolumn{15}{c}{OntoNotes-WB} \\
    \midrule
    & &&&&&&& \multicolumn{1}{l}{\rotatebox{90}{eDen}} & \multicolumn{1}{l}{\rotatebox{90}{oDen}} & \multicolumn{1}{l}{\rotatebox{90}{sLen}} & \multicolumn{1}{l}{\rotatebox{90}{eCon}} & \multicolumn{1}{l}{\rotatebox{90}{eFre}} & \multicolumn{1}{l}{\rotatebox{90}{tCon}} & \multicolumn{1}{l}{\rotatebox{90}{tFre}} & \multicolumn{1}{l}{\rotatebox{90}{eLen}} &
     &&&&&&& \multicolumn{1}{l}{\rotatebox{90}{eDen}} & \multicolumn{1}{l}{\rotatebox{90}{oDen}} & \multicolumn{1}{l}{\rotatebox{90}{sLen}} & \multicolumn{1}{l}{\rotatebox{90}{eCon}} & \multicolumn{1}{l}{\rotatebox{90}{eFre}} & \multicolumn{1}{l}{\rotatebox{90}{tCon}} & \multicolumn{1}{l}{\rotatebox{90}{tFre}} & \multicolumn{1}{l}{\rotatebox{90}{eLen}} &
    &&&&&&& \multicolumn{1}{l}{\rotatebox{90}{eDen}} & \multicolumn{1}{l}{\rotatebox{90}{oDen}} & \multicolumn{1}{l}{\rotatebox{90}{sLen}} & \multicolumn{1}{l}{\rotatebox{90}{eCon}} & \multicolumn{1}{l}{\rotatebox{90}{eFre}} & \multicolumn{1}{l}{\rotatebox{90}{tCon}} & \multicolumn{1}{l}{\rotatebox{90}{tFre}} & \multicolumn{1}{l}{\rotatebox{90}{eLen}} &
    &&&&&&& \multicolumn{1}{l}{\rotatebox{90}{eDen}} & \multicolumn{1}{l}{\rotatebox{90}{oDen}} & \multicolumn{1}{l}{\rotatebox{90}{sLen}} & \multicolumn{1}{l}{\rotatebox{90}{eCon}} & \multicolumn{1}{l}{\rotatebox{90}{eFre}} & \multicolumn{1}{l}{\rotatebox{90}{tCon}} & \multicolumn{1}{l}{\rotatebox{90}{tFre}} & \multicolumn{1}{l}{\rotatebox{90}{eLen}} &
    &&&&&&& \multicolumn{1}{l}{\rotatebox{90}{eDen}} & \multicolumn{1}{l}{\rotatebox{90}{oDen}} & \multicolumn{1}{l}{\rotatebox{90}{sLen}} & \multicolumn{1}{l}{\rotatebox{90}{eCon}} & \multicolumn{1}{l}{\rotatebox{90}{eFre}} & \multicolumn{1}{l}{\rotatebox{90}{tCon}} & \multicolumn{1}{l}{\rotatebox{90}{tFre}} & \multicolumn{1}{l}{\rotatebox{90}{eLen}} &
    &&&&&&& \multicolumn{1}{l}{\rotatebox{90}{eDen}} & \multicolumn{1}{l}{\rotatebox{90}{oDen}} & \multicolumn{1}{l}{\rotatebox{90}{sLen}} & \multicolumn{1}{l}{\rotatebox{90}{eCon}} & \multicolumn{1}{l}{\rotatebox{90}{eFre}} & \multicolumn{1}{l}{\rotatebox{90}{tCon}} & \multicolumn{1}{l}{\rotatebox{90}{tFre}} & \multicolumn{1}{l}{\rotatebox{90}{eLen}} \\
\midrule 
    \multicolumn{1}{l}{Overall F1} & 
     \multicolumn{15}{c}{M1: 91.11} &
     \multicolumn{15}{c}{M1: 47.77} &
     \multicolumn{15}{c}{M1: 86.90}  &
     \multicolumn{15}{c}{M1: 81.03} &
     \multicolumn{15}{c}{M1: 89.64}&
     \multicolumn{15}{c}{M1: 66.35} \\
    \cmidrule(r){1-1}\cmidrule(lr){2-16}\cmidrule(lr){17-33}\cmidrule(lr){34-49}\cmidrule(lr){49-63}\cmidrule(lr){64-79}\cmidrule(lr){80-95} 

    & \multicolumn{15}{l}{\multirow{12}[2]{*}{\includegraphics[scale=0.39]{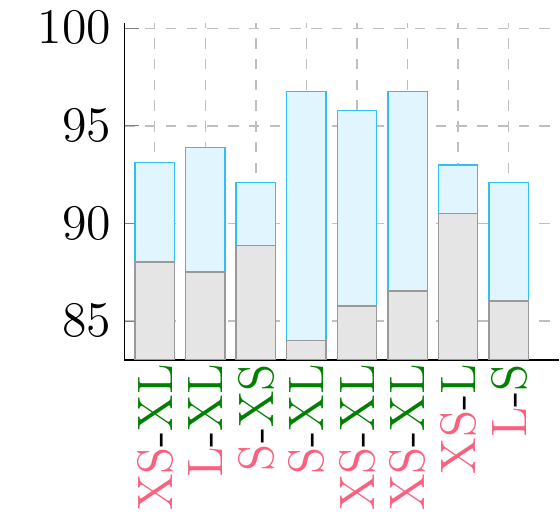}}}              
    & \multicolumn{15}{l}{\multirow{12}[2]{*}{\includegraphics[scale=0.39]{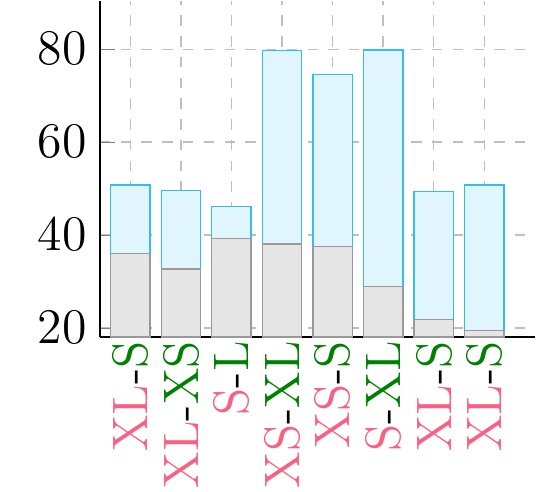}}}              
    & \multicolumn{15}{l}{\multirow{12}[2]{*}{\includegraphics[scale=0.39]{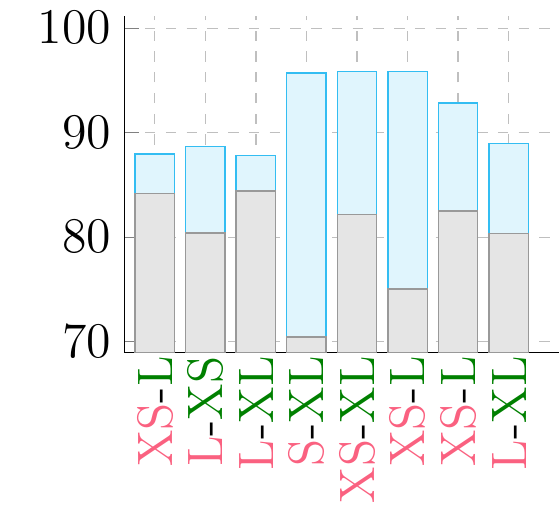}}}              
    & \multicolumn{15}{l}{\multirow{12}[2]{*}{\includegraphics[scale=0.39]{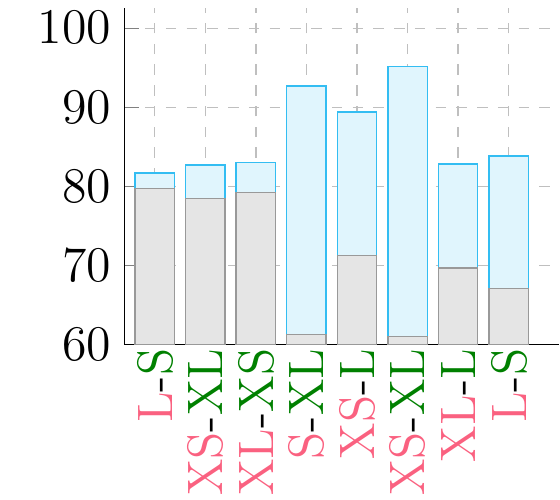}}} 
    &\multicolumn{15}{l}{\multirow{12}[2]{*}{\includegraphics[scale=0.39]{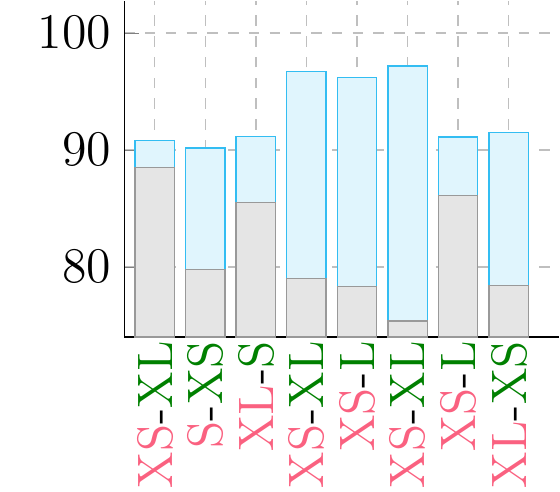}}} 
    &\multicolumn{15}{l}{\multirow{12}[2]{*}{\includegraphics[scale=0.39]{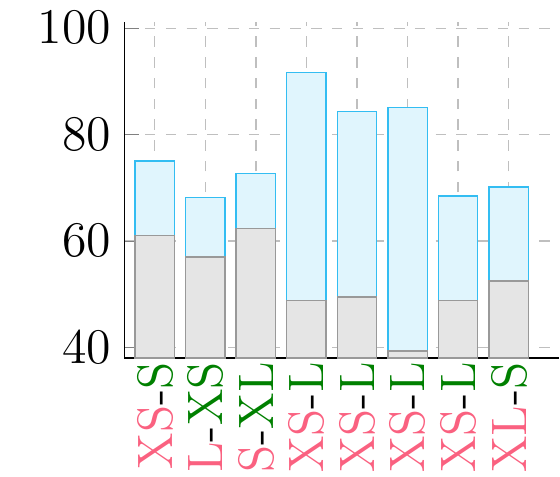}}} 
    \\ \\ \\ \\ \\ \\
    \multicolumn{1}{l}{M1: \textit{CbertWnoneLstmMlp}} & 
    \\  \\
    \multicolumn{1}{c}{\textbf{Self-diagnosis}} & 
    \\ \\ \\ \\

\midrule 
      \multicolumn{1}{l}{Overall F1} & 
     \multicolumn{15}{c}{M1: 90.48; M2: 88.05} &
     \multicolumn{15}{c}{M1: 40.61; M2: 32.84} &
     \multicolumn{15}{c}{M1: 85.39; M2: 81.09}  &
     \multicolumn{15}{c}{M1: 76.04; M2: 70.00} &
     \multicolumn{15}{c}{M1: 86.78; M2: 84.07}&
     \multicolumn{15}{c}{M1: 60.17; M2: 56.61}\\
    \cmidrule(r){1-1}\cmidrule(lr){2-16}\cmidrule(lr){17-33}\cmidrule(lr){34-49}\cmidrule(lr){49-63}\cmidrule(lr){64-79}\cmidrule(lr){80-95} 

    & \multicolumn{15}{c}{\multirow{12}[2]{*}{\includegraphics[scale=0.39]{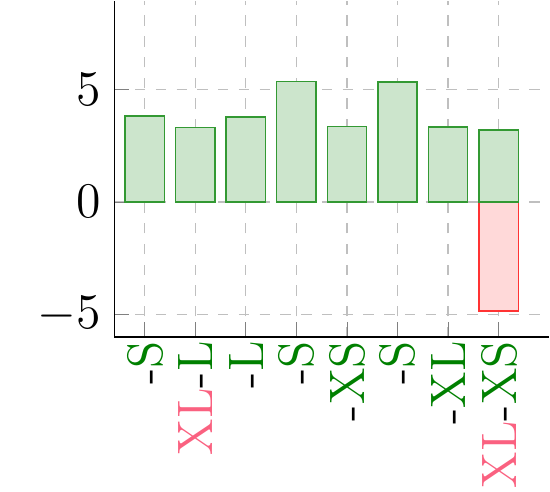}}}              
    & \multicolumn{15}{c}{\multirow{12}[2]{*}{\includegraphics[scale=0.39]{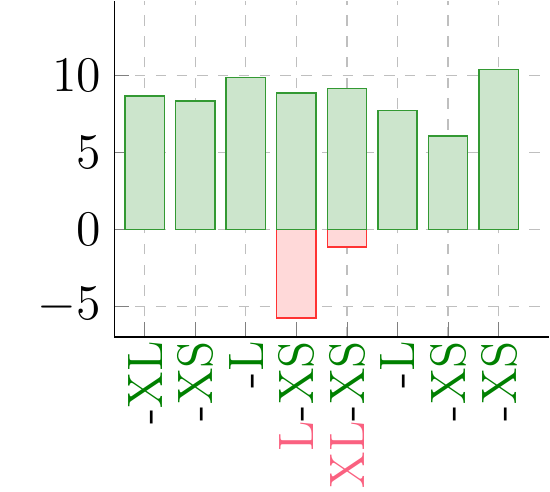}}}              
    & \multicolumn{15}{c}{\multirow{12}[2]{*}{\includegraphics[scale=0.39]{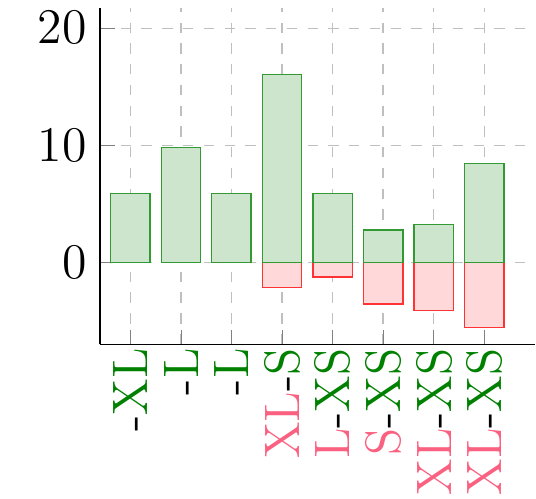}}}              
    & \multicolumn{15}{c}{\multirow{12}[2]{*}{\includegraphics[scale=0.39]{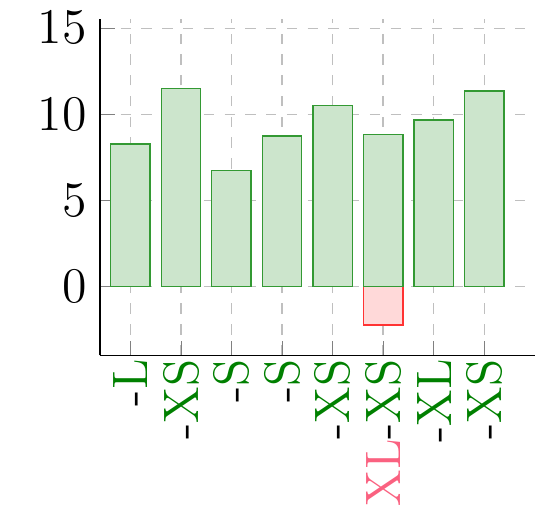}}} 
    &\multicolumn{15}{c}{\multirow{12}[2]{*}{\includegraphics[scale=0.39]{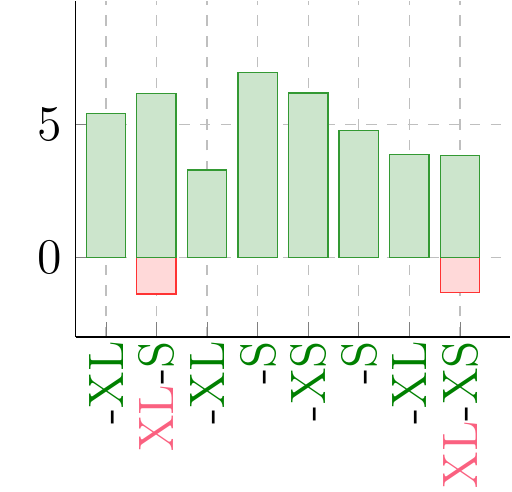}}} 
    &\multicolumn{15}{c}{\multirow{12}[2]{*}{\includegraphics[scale=0.39]{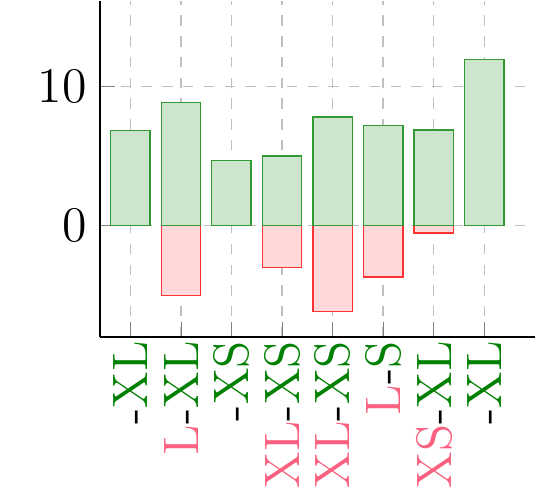}}} 
    \\ \\ \\ \\
     \multicolumn{1}{l}{M1: \textit{CcnnWgloveLstmCrf}} &  
    \\ \\ 
    \multicolumn{1}{l}{M2: \textit{CcnnWgloveLstmMlp}} & 
    \\  \\
    \multicolumn{1}{c}{\textbf{Comparative diagnosis}} & 
    \\ \\ \\ \\ 

    \bottomrule
    \end{tabular}%
    \vspace{-5pt}
  \caption{\textit{Self-diagnosis}, and \textit{Comparative diagnosis} (Sec.~\ref{sec:bucket-wise-eq}) of different NER systems. M1 and M2 denote two models.
   We classify the attribute values into four categories: extra-small (XS), small (S), large (L) and extra-large (XL).
  In the \textit{self-diagnosis} histogram, \textcolor{aogreen}{green} (\textcolor{brinkpink}{red}) $x$ ticklabels represents the bucket value of a specific attribute on which system achieved best (worst) performance.
 Gray bins represent worst performance while  blue bins denote the gap between best and worst performance.
 In the \textit{comparative diagnosis} histogram, \textcolor{aogreen}{green} (\textcolor{brinkpink}{red}) $x$ ticklabels represents the bucket value of a specific attribute on which system M1 surpasses (under-performs) M2 by the largest margin that is illustrated by a \textcolor{aogreen}{green} (\textcolor{brinkpink}{red}) bin.
 }
  \label{tab:bucket-wise}%
\end{table*}%

\subsection{Exp-III: Bucket-wise Analysis} \label{sec:bucket-wise}
Bucket-wise analysis aims to identify the buckets that satisfy some specific constraints. In this paper, we present two flavors of diagnostic: \textit{self-diagnosis} and \textit{comparative diagnosis}.

\subsubsection{Approach}
\label{sec:bucket-wise-eq}

\paragraph{Self-diagnosis}
Given a model $M_1$ and a specific evaluation attribute (e.g., \texttt{eLen}), self-diagnosis selects the buckets in which test samples have achieved the highest and lowest performance (F1 score). Intuitively, this operation can help us diagnose under which conditions a particular model performs well or poorly:
$
    \mathrm{SelfDiag}(M_1) = \mathrm{argFunc}_k \mathcal{T}[M_1,j,k]
$
where $\mathrm{argFunc}$ can be instantiated as $\mathrm{argMax}$ and $\mathrm{argMin}$.

\paragraph{Comparative diagnosis}
Given two models $M_1$, $M_2$ and an attribute, comparative diagnosis aims to select buckets in which the performance gap between the two systems
achieve the highest and lowest values.
This method can indicate under which conditions a particular system may have a relative advantage over another system:
$
    \mathrm{CoDiag}(M_1,M_2) = \mathrm{argFunc}_k (\mathcal{T}[M_1,j,k]-\mathcal{T}[M_2,j,k])
$

\paragraph{Significance Tests}
We test for statistical significance at $p= 0.05$ with Wilcoxon's signed-rank test \citep{wilcoxon1970critical}. The null hypothesis is that, given a specific attribute value (e.g. long entities \texttt{eLen:XL}), two different models have the same expected performance.\footnote{We opt for Wilcoxon's signed-rank test instead of Friedman's test because the diagnosis (self- or comparative diagnosis) only has two group samples while the Friedman's test requires more than two groups \cite{zimmerman1993relative}.}

\subsubsection{Self-Diagnosis}
\label{sec:self-diagnosis-analysis}
\paragraph{BERT}
The first row in Tab.~\ref{tab:bucket-wise} illustrates the \textit{self-diagnosis} of the model \textit{CbertWnonelstmMlp}.
The \textcolor{aogreen}{green} (\textcolor{brinkpink}{red}) $x$ ticklabels represent the bucket value of a specific attribute on which system has achieved best (worst) performance.
 Gray bins represent worst performance while  blue bins denote the gap between best and worst performance. 

We observe that large performance gaps (tall blue bins) commonly occur for the attributes label consistency and entity frequency, and the worst performance on these attributes was obtained on buckets with low consistency (\texttt{eCon, tCon}:\texttt{\textcolor{brinkpink}{XS/S}}) and low entity frequency (\texttt{eFre}:\texttt{\textcolor{brinkpink}{S}}).

We conduct significance testing  on the worst and best performances\footnote{We restarted the BERT-based system twice on six datasets, and we got 12 best and 12 worst F1 scores for a given attribute. } of \texttt{eCon} ($1.7\times 10^{-8}$), \texttt{tCon} ($2.3\times 10^{-7}$) and \texttt{eFre} ($1.2\times 10^{-5}$) respectively, and they all passed with $p<0.05$. 
This reveals that it is still challenging for contextualized pre-trained NER systems to handle entities with lower \textit{label consistency} and lower \textit{entity frequency}.

\subsubsection{Comparative Diagnosis}
We highlight major observations and include more analysis in the appendix.

\paragraph{CRF v.s. MLP}
\label{sec:comp-diagnosis-crf}
The benefits of using CRF on the sentence with high entity density (\texttt{eDen}:\texttt{\textcolor{aogreen}{XL}}) are remarkably stable, and improvement can be seen in all datasets ($p=1.8\times10^{-5}<0.05$).
Similarly, based on attribute-wise metric $\zeta$ in Fig.~\ref{fig:spider}(a), we find \textit{label consistency} (\texttt{eCon}, \texttt{tCon}) is a major factor for the choices of CRF and MLP layers:

1) Introducing a CRF achieves larger improvements on long entities once the dataset has a lower \textit{label consistency} (e.g. $\zeta_{\texttt{eCon},\texttt{tCon}}$(\texttt{WNUT}), $\zeta_{\texttt{eCon},\texttt{tCon}}$(\texttt{WB}), and $\zeta_{\texttt{eCon},\texttt{tCon}}$(\texttt{BC}) are lowest).
We conduct the significance testing on CRF and MLP systems with respect to the long entities on these three datasets\footnote{We restarted the CRF and MLP systems on \texttt{WNUT}, \texttt{WB}, and \texttt{BC} for $5$ times, and we got $3\times5=15$ F1 scores on CRF and MLP systems respectively.} (\texttt{WNUT}, \texttt{WB}, and \texttt{BC}), and the result indicates that the performance of the CRF and MLP systems are significantly different on long entity bucket ($p=6.5\times10^{-4}<0.05$).
2) by contrast, if a dataset has a higher \textit{label consistency} ($\zeta_{\texttt{eCon},\texttt{tCon}}$(\texttt{CoNLL}),$\zeta_{\texttt{eCon},\texttt{tCon}}$(\texttt{BN}), $\zeta_{\texttt{eCon},\texttt{tCon}}$(\texttt{MZ}) are highest), using the CRF layer does not exhibit significant gains (even worse than models without CRF) on longer entities (\texttt{eLen:XL}). We do significance testing like 1), and $p=5.1\times10^{-3}<0.05$.

\section{Application: Well-grounded Model Improvement}
The purpose of interpretable evaluation and analysis is to provide more evidence for us to rethink current learning models and move forward.
In what follows, we choose a piece of evidence observed from the above analysis and attempt to present one simple solution to improve the model.
From Sec.~\ref{sec:attribute-wise} and Sec.~\ref{sec:self-diagnosis-analysis}, we know that \textit{label consistency} is a decisive factor, and even more advanced models (e.g., \texttt{BERT}, \texttt{Flair}) fail to consistently predict entities with low \textit{label consistency}.
An intuitive idea to disambiguate these entities is using more contextual information. To this end, we shift the setting of traditional sentence-level training and testing to use larger context, and investigate this change's effectiveness.

\subsection{Experimental Setting}
We choose \textit{CbertWnoneLstmMlp} as a base model, which will be trained under different numbers ($K = 1,2,3,4,5,6,10$) of contextual sentences on all six datasets respectively. For example, $K =2$ represents that each training sample is constructed by concatenating two consecutive sentences from the original dataset ($K =1$).

\begin{figure}[ht]
    \centering \footnotesize
     \subfloat[CoNLL]{
    \includegraphics[width=0.3\linewidth]{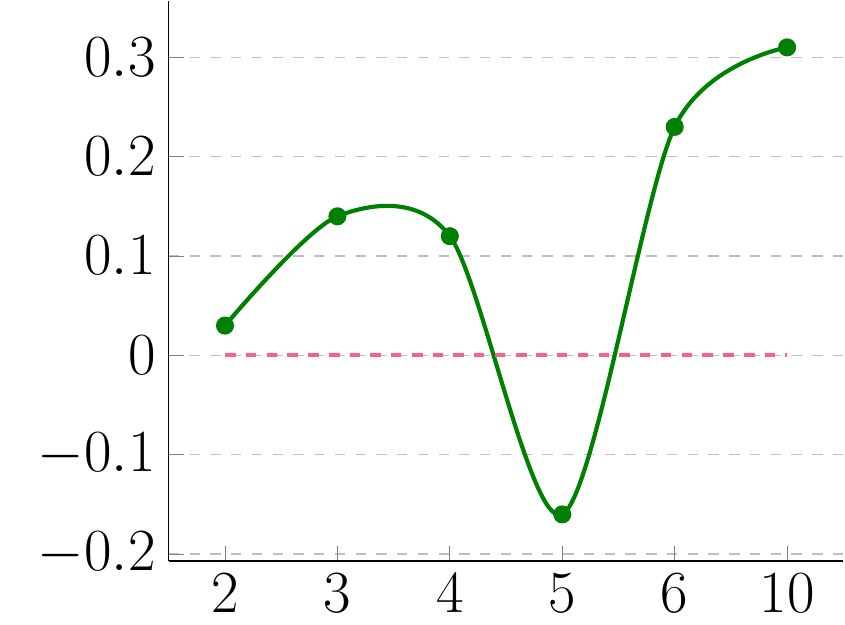} 
    }  \hspace{0.01em}  
    \subfloat[BN]{
    \includegraphics[width=0.31\linewidth]{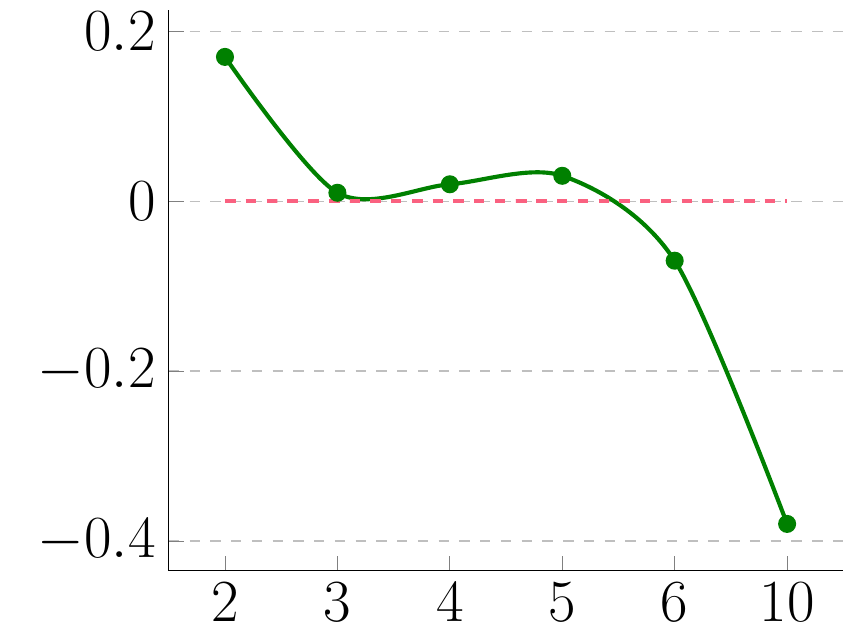} 
    }  \hspace{0.01em} 
    \subfloat[MZ]{
    \includegraphics[width=0.3\linewidth]{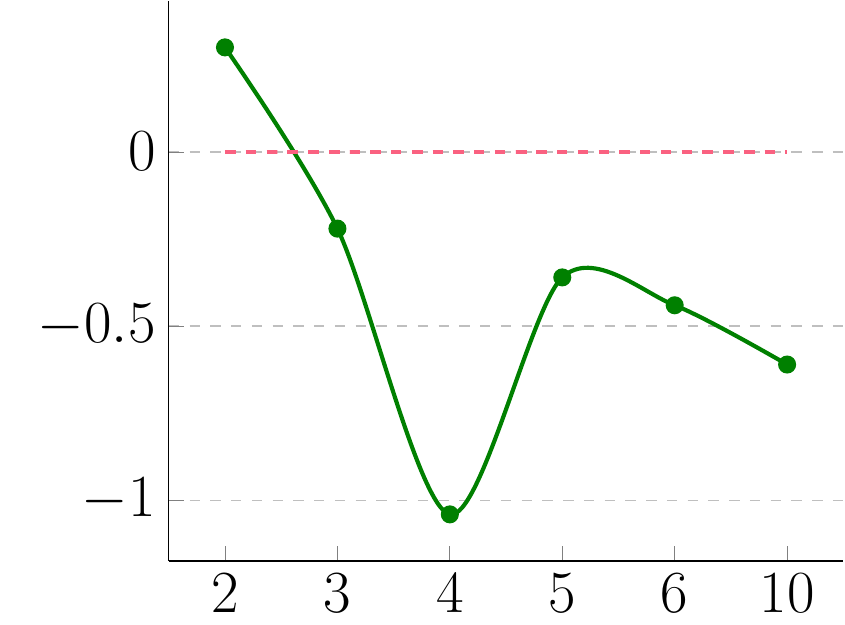}
    }        
    \\
    \vspace{-10pt}
    \subfloat[BC]{
    \includegraphics[width=0.3\linewidth]{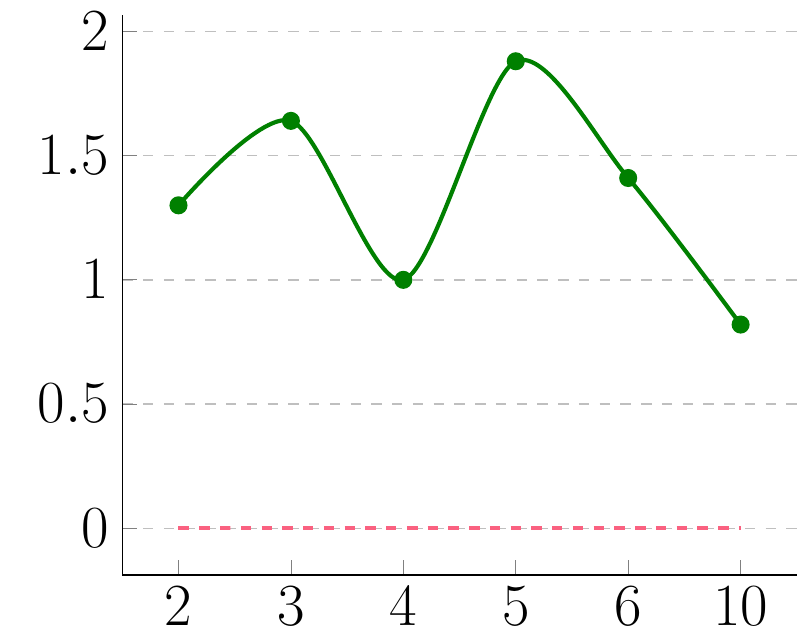}
    } \hspace{0.01em}   
    \subfloat[WB]{
    \includegraphics[width=0.3\linewidth]{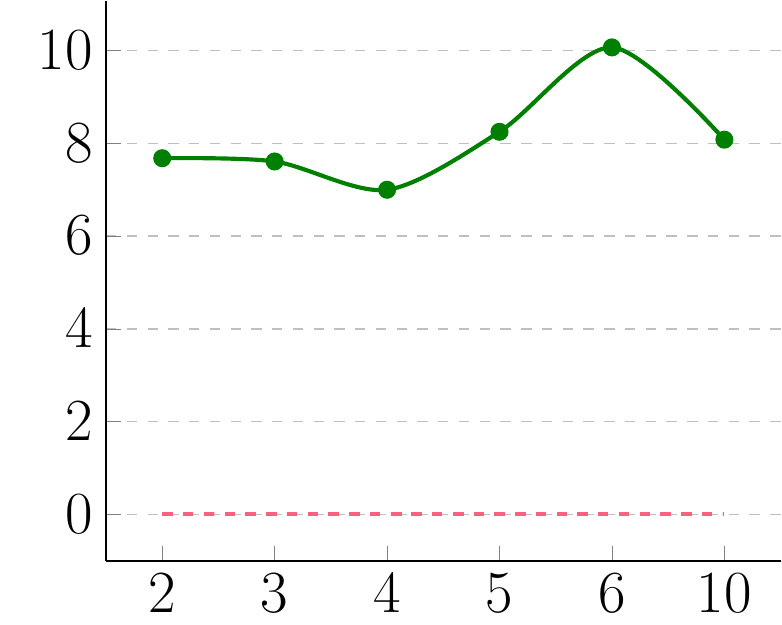}
    }\hspace{0.01em}
    \subfloat[WNUT]{
    \includegraphics[width=0.3\linewidth]{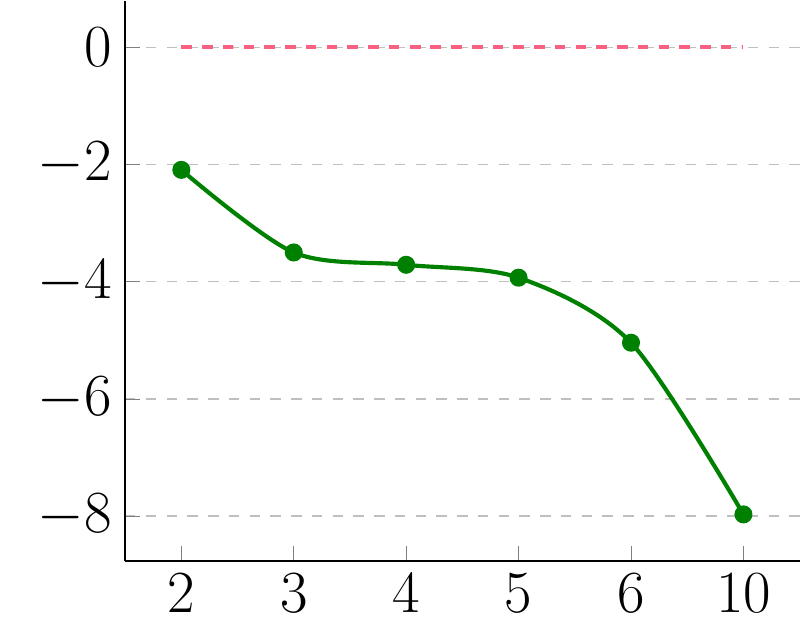}
    } 
    \vspace{-5pt}
    \caption{Illustration of the improvement achieved by the larger context method with different sizes ($K$) on different datasets. The part above the red suggests the improvement brought by the corresponding value of $K$.} 
    \label{fig:larger_context_line}
\end{figure}

\begin{figure}[htb]
    \centering \footnotesize
     \subfloat[eCon]{
    \includegraphics[width=0.23\linewidth]{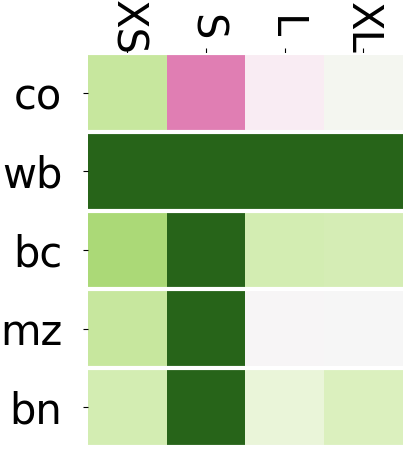}
    }    \hspace{-0.1em}
    \subfloat[tCon]{
    \includegraphics[width=0.18\linewidth]{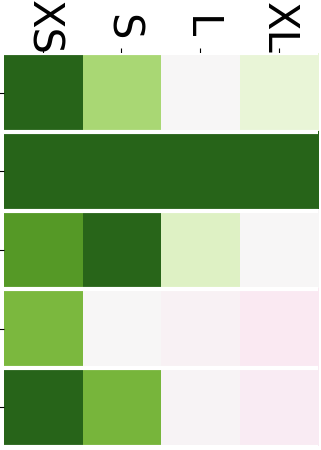}
    } \hspace{-0.1em}
    \subfloat[eFre]{
    \includegraphics[width=0.18\linewidth]{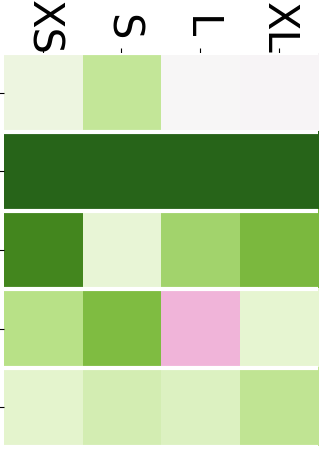}
    }  \hspace{-0.1em}
    \subfloat[tFre]{
    \includegraphics[width=0.26\linewidth]{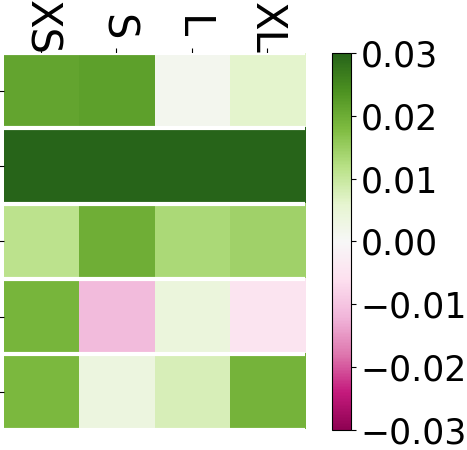} 
    } \\
    \vspace{-8pt}
    \subfloat[eDen]{
    \includegraphics[width=0.23\linewidth]{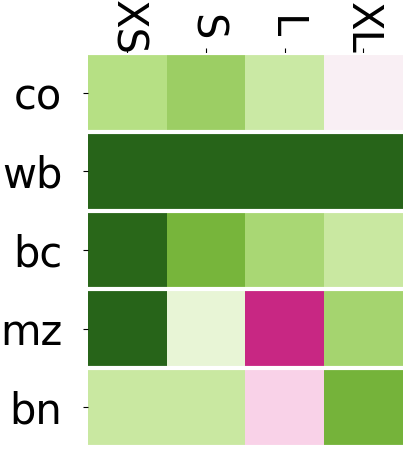}
    }  
    \hspace{-0.1em}
    \subfloat[eLen]{
    \includegraphics[width=0.18\linewidth]{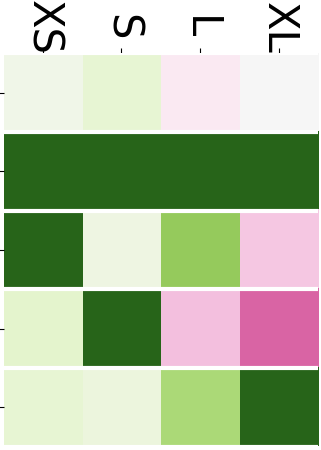}
    } \hspace{-0.1em}
    \subfloat[sLen]{
    \includegraphics[width=0.18\linewidth]{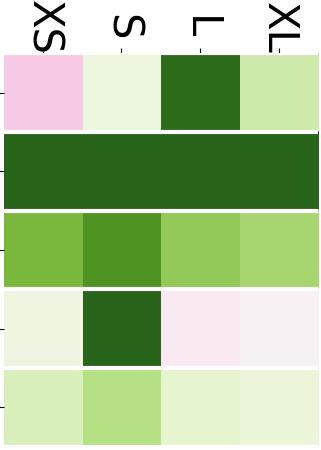}\hspace{-0.1em}
    }
    \subfloat[oDen]{
    \includegraphics[width=0.26\linewidth]{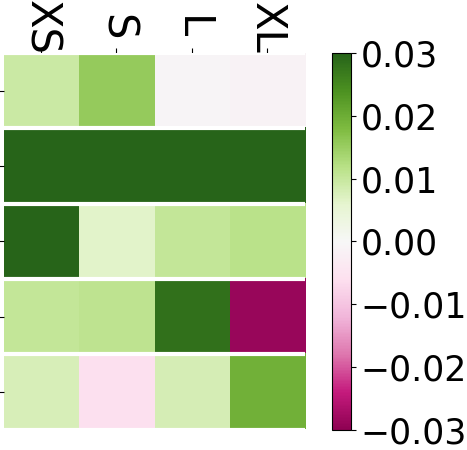}
    } 
    \vspace{-7pt}
    \caption{The relative increase of the larger-context method on five datasets\footnotemark~based on eight evaluation attributes.    ``\texttt{Co}'' represents the dataset {CoNLL-2003} while ``\texttt{wb}'', ``\texttt{bc}'', ``\texttt{mz}'', ``\texttt{bn}'' denote different domains from the {OntoNotes}. } 
    \label{fig:larger_context_heat}
\end{figure}
\footnotetext{We leave \texttt{WNUT} out due to its worse performance.}

\subsection{Results and Analysis}
\paragraph{Results} As presented in Fig.~\ref{fig:larger_context_line}, the green line describes the relative improvement of the larger context method compared with the vanilla model ($K=1$) with different numbers of context sentences $K = 2,\cdots,10$.
In detail, we observed:

\label{sec:app-longcontex-wb}
1) For most of the datasets (except ``\texttt{WNUT}''), the performance increases as more context sentences are introduced.
2) Surprisingly, we achieved a \textbf{10.07 improvement} ($66.35$ vs. $76.42$, significance testing result\footnote{We restart the system on  \texttt{WB} with $K=1$ and $K=6$ setting for $10$ times respectively.}: $p=5.1\times10^{-3}<0.05$) F1 score on dataset ``\texttt{WB}'', with such a simple larger-context training method. 
3) There is no gain on ``\texttt{WNUT}'', and the reason can be attributed to lack of dependency between samples, which are collected from \textit{Twitter}\footnote{https://twitter.com/} where each sentence is relatively independent with the another.

\paragraph{Analysis using Multi-dimensional Evaluation}
To probe into where the gain afforded by larger context comes from, we use our proposed evaluation attributes to conduct a fine-grained investigation, aiming to answer the question: \textit{how does this method influence different datasets' performance seen from different attributes?} (e.g.,  \textit{label consistency} of entity, \texttt{eCon}).
As expressed in Fig.~\ref{fig:larger_context_heat}, the value of each unit in the heat maps denotes the relative increase achieved by the larger-context method.
Intuitively, a darker green area implies more significant improvement while a darker red unit suggests larger-context leads to worse performance. 

Different evaluation attributes allow us to understand the source of improvement from diverse perspectives:
1) in terms of \textit{label consistency} (\texttt{eCon}, \texttt{tCon}), test entities with lower \textit{label consistency} will achieve larger improvements with the help of more contextual sentences. Importantly, from Fig.~\ref{fig:larger_context_heat} we can see this observation holds true for all datasets.
2) in terms of entity length (\texttt{eLen}),  larger-context information has no advantage in dealing with longer entities (\texttt{L}, \texttt{XL}). For example, in the three of five datasets, more contextual sentences lead to worse performance on longer test entities.

\vspace{-6pt}
\section{Discussion} \label{sec:conclusion}
This paper has provided a framework where we can covert our understanding of the NER task (i.e., which attributes matter for the current task?) into interpretable evaluation aspects, and define axes through which we can apply them to acquire insights and make model improvements.
This is just a first step towards the goal of fully-automated interpretable evaluation, and applications to new attributes and tasks beyond NER are promising future directions.

\section*{Acknowledgments}
We would like to thank the anonymous reviewers for their valuable comments.
This material is based on research sponsored by the Air Force Research
Laboratory under agreement number FA8750-19-2-0200. The U.S. Government
is authorized to reproduce and distribute reprints for Governmental
purposes notwithstanding any copyright notation thereon. The views and
conclusions contained herein are those of the authors and should not be
interpreted as necessarily representing the official policies or
endorsements, either expressed or implied, of the Air Force Research
Laboratory or the U.S. Government.

\bibliography{emnlp.bib}
\bibliographystyle{acl_natbib}

\appendix

\section{Models Description}
Tab.~\ref{tab:allmodels} shows the evaluated models in this paper, mainly in terms of four aspects:
1) character/subword-sensitive encoder: \texttt{ELMo}, \texttt{Flair}, \texttt{BERT}
2) additional word embeddings: \texttt{GloVe};
3) sentence-level encoders: \texttt{LSTM}, \texttt{CNN};
4) decoders: \texttt{MLP} or \texttt{CRF}.

For example,
1) "\textit{CnonWrandlstmCrf}" is a model with no character features, with randomly initialized embeddings, and the sentence encoder is LSTM and the decoder is CRF. 
2) "\textit{CbertWnoneLstmMlp}" is a model that concatenates the representations from BERT and GloVe as a subword-sensitive encoder. Then the concatenation will be fed into an MLP layer, predicting a label over all classes.
3) "\textit{CelmWgloveLstmCrf}" is a model that concatenates the representations from ELMo and GloVe as a subword-sensitive encoder.
Then the concatenation will be fed into an LSTM layer, followed by the CRF layer.

\section{Bucketing Interval Strategy}
In this section, we will illustrate the bucketing interval with respect to attribute. 
We divide the range of attribute values into $m$ discrete parts. For a given attribute, the number of entities covered by an attribute value is various. For example, \texttt{oDen=0} covered nearly half of the entity in the test set for \textit{OOV density}; for \textit{label consistency}, \texttt{eCon=0} and \texttt{eCon=1} each occupy a large part of the test entities. We customize the interval method for each attribute in accordance with its own characteristics. 

1) \textbf{Label consistency (\texttt{eCon}, \texttt{tCon})}: first, we divide the entities in the test set with attribute values  $\phi_{eCon}(\mathbf{x})=0$ and $\phi_{eCon}(\mathbf{x})=1$ into the first bucket ($\mathcal{E}_1^{te}$) and last bucket $\mathcal{E}_m^{te}$, respectively; then, divide the remaining entities equally into $m-2$ buckets. The bucketing interval strategy of \texttt{eCon} is suitable for \texttt{tCon}.

2) \textbf{Frequency (\texttt{eFre}, \texttt{tFre}) and OOV density (\texttt{oDen})}: first, we divide the entities in test set with attribute value $\phi_{eFre}(\mathbf{x})=0$ into the first bucket ($\mathcal{E}_1^{te}$); then, divide the remaining entities equally into $m-1$ buckets. The bucketing interval strategy of \texttt{eFre} is suitable for \texttt{tFre} and \texttt{oDen}.

3) \textbf{Sentence length (\texttt{sLen}) and entity density (\texttt{eDen})}: we divide the test entities equally into $m$ buckets.

4) \textbf{Entity length (\texttt{eLen})}: a small $m$ is suitable for \textit{entity length}, because of a few attribute values (generally, the entity length is rarely greater than 6). In this paper, we put the entities in the test set with lengths of 1, 2, 3, and $\geq 4$ into four buckets, respectively.

\renewcommand\tabcolsep{0.6pt}

\renewcommand\tabcolsep{0.9pt}
\renewcommand\arraystretch{0.73}  
\begin{table*}[!htb]
  \centering \tiny
    \begin{tabular}{ccccccc ccccccc ccccccc ccccccc ccccccc ccccccc ccccccc ccccccc ccccccc ccccccc ccccccc ccccccc ccccccc cccc} 
    \toprule
          & \multicolumn{15}{c}{CoNLL03}                           & \multicolumn{15}{c}{WNUT16}                            & \multicolumn{15}{c}{OntoNotes-MZ}                            & \multicolumn{15}{c}{OntoNotes-BC}  &
          \multicolumn{15}{c}{OntoNotes-BN}  &
          \multicolumn{15}{c}{OntoNotes-WB} \\
    \midrule
    & &&&&&&& \multicolumn{1}{l}{\rotatebox{90}{eDen}} & \multicolumn{1}{l}{\rotatebox{90}{oDen}} & \multicolumn{1}{l}{\rotatebox{90}{sLen}} & \multicolumn{1}{l}{\rotatebox{90}{eCon}} & \multicolumn{1}{l}{\rotatebox{90}{eFre}} & \multicolumn{1}{l}{\rotatebox{90}{tCon}} & \multicolumn{1}{l}{\rotatebox{90}{tFre}} & \multicolumn{1}{l}{\rotatebox{90}{eLen}} &
     &&&&&&& \multicolumn{1}{l}{\rotatebox{90}{eDen}} & \multicolumn{1}{l}{\rotatebox{90}{oDen}} & \multicolumn{1}{l}{\rotatebox{90}{sLen}} & \multicolumn{1}{l}{\rotatebox{90}{eCon}} & \multicolumn{1}{l}{\rotatebox{90}{eFre}} & \multicolumn{1}{l}{\rotatebox{90}{tCon}} & \multicolumn{1}{l}{\rotatebox{90}{tFre}} & \multicolumn{1}{l}{\rotatebox{90}{eLen}} &
    &&&&&&& \multicolumn{1}{l}{\rotatebox{90}{eDen}} & \multicolumn{1}{l}{\rotatebox{90}{oDen}} & \multicolumn{1}{l}{\rotatebox{90}{sLen}} & \multicolumn{1}{l}{\rotatebox{90}{eCon}} & \multicolumn{1}{l}{\rotatebox{90}{eFre}} & \multicolumn{1}{l}{\rotatebox{90}{tCon}} & \multicolumn{1}{l}{\rotatebox{90}{tFre}} & \multicolumn{1}{l}{\rotatebox{90}{eLen}} &
    &&&&&&& \multicolumn{1}{l}{\rotatebox{90}{eDen}} & \multicolumn{1}{l}{\rotatebox{90}{oDen}} & \multicolumn{1}{l}{\rotatebox{90}{sLen}} & \multicolumn{1}{l}{\rotatebox{90}{eCon}} & \multicolumn{1}{l}{\rotatebox{90}{eFre}} & \multicolumn{1}{l}{\rotatebox{90}{tCon}} & \multicolumn{1}{l}{\rotatebox{90}{tFre}} & \multicolumn{1}{l}{\rotatebox{90}{eLen}} &
    &&&&&&& \multicolumn{1}{l}{\rotatebox{90}{eDen}} & \multicolumn{1}{l}{\rotatebox{90}{oDen}} & \multicolumn{1}{l}{\rotatebox{90}{sLen}} & \multicolumn{1}{l}{\rotatebox{90}{eCon}} & \multicolumn{1}{l}{\rotatebox{90}{eFre}} & \multicolumn{1}{l}{\rotatebox{90}{tCon}} & \multicolumn{1}{l}{\rotatebox{90}{tFre}} & \multicolumn{1}{l}{\rotatebox{90}{eLen}} &
    &&&&&&& \multicolumn{1}{l}{\rotatebox{90}{eDen}} & \multicolumn{1}{l}{\rotatebox{90}{oDen}} & \multicolumn{1}{l}{\rotatebox{90}{sLen}} & \multicolumn{1}{l}{\rotatebox{90}{eCon}} & \multicolumn{1}{l}{\rotatebox{90}{eFre}} & \multicolumn{1}{l}{\rotatebox{90}{tCon}} & \multicolumn{1}{l}{\rotatebox{90}{tFre}} & \multicolumn{1}{l}{\rotatebox{90}{eLen}}  \\

\midrule 
    \multicolumn{1}{l}{Overall F1} & 
     \multicolumn{15}{c}{M1: 90.48; M2: 90.14} &
     \multicolumn{15}{c}{M1: 40.61; M2: 36.21} &
     \multicolumn{15}{c}{M1: 85.39; M2: 88.10}  &
     \multicolumn{15}{c}{M1: 76.04; M2: 76.74} &
     \multicolumn{15}{c}{M1: 86.78; M2: 86.42}&
     \multicolumn{15}{c}{M1: 60.17; M2: 49.10}\\
    \cmidrule(r){1-1}\cmidrule(lr){2-16}\cmidrule(lr){17-33}\cmidrule(lr){34-49}\cmidrule(lr){49-63}\cmidrule(lr){64-79}\cmidrule(lr){80-95} 

    & \multicolumn{15}{c}{\multirow{12}[2]{*}{\includegraphics[scale=0.39]{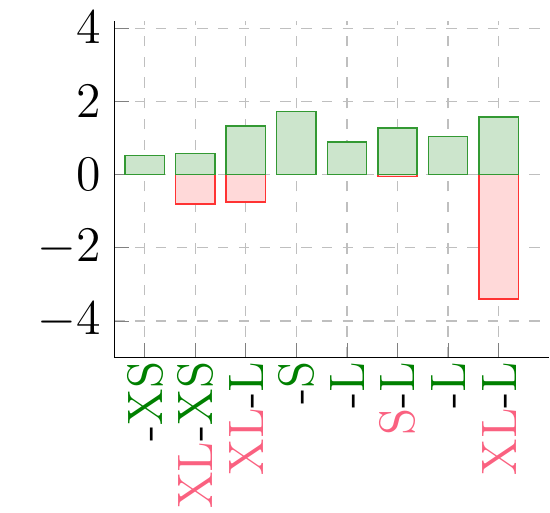}}}              
    & \multicolumn{15}{c}{\multirow{12}[2]{*}{\includegraphics[scale=0.39]{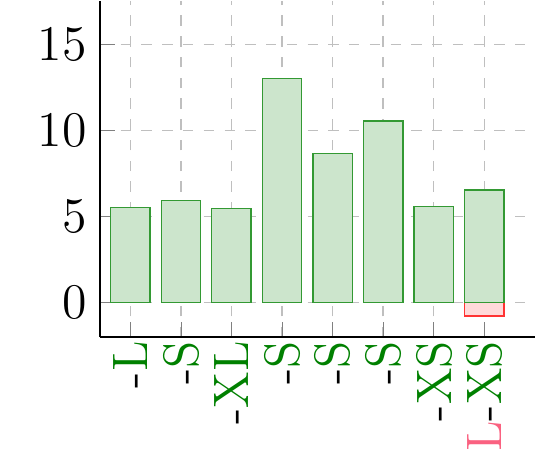}}}              
    & \multicolumn{15}{c}{\multirow{12}[2]{*}{\includegraphics[scale=0.39]{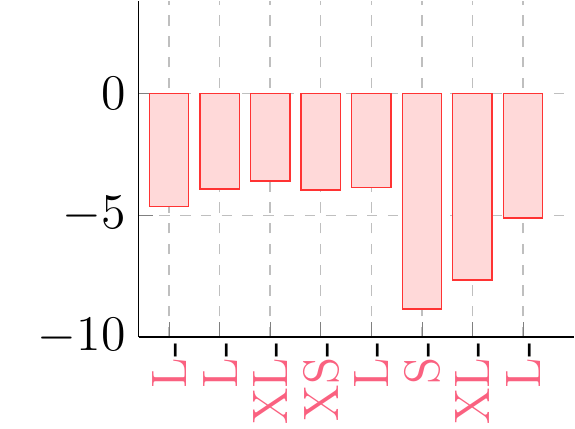}}}              
    & \multicolumn{15}{c}{\multirow{12}[2]{*}{\includegraphics[scale=0.39]{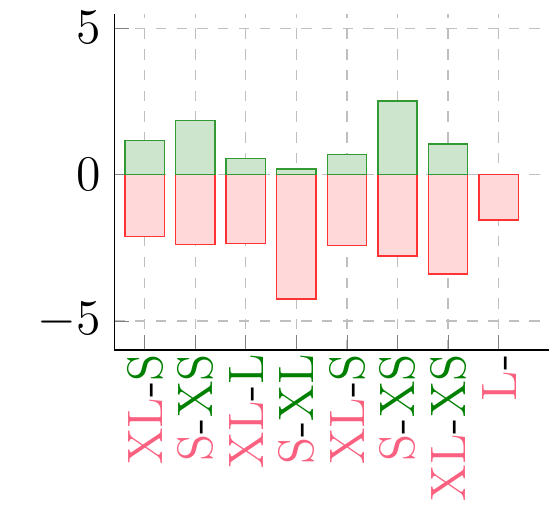}}} 
    &\multicolumn{15}{c}{\multirow{12}[2]{*}{\includegraphics[scale=0.39]{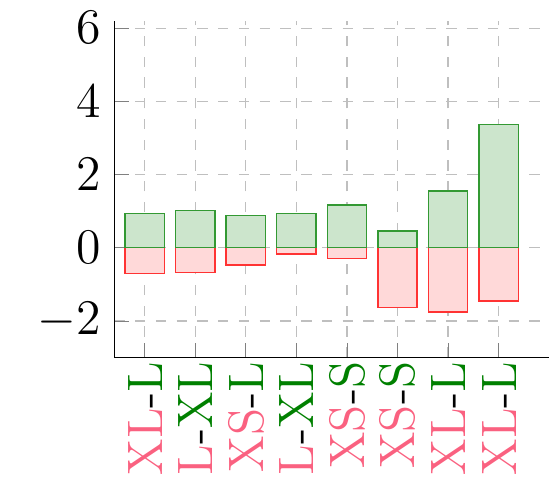}}} 
    &\multicolumn{15}{c}{\multirow{12}[2]{*}{\includegraphics[scale=0.39]{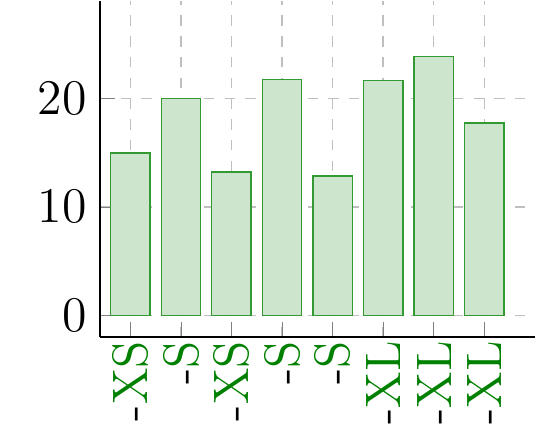}}} 
    \\ \\ \\ \\
     \multicolumn{1}{l}{M1: \textit{CcnnWgloveLstmCrf}} &  
    \\ \\ 
    \multicolumn{1}{l}{M2: \textit{CcnnWgloveCnnCrf}} & 
    \\  \\
    \multicolumn{1}{c}{\textcolor{brinkpink}{Comparative diagnosis}} & 
    \\ \\ \\ \\

\midrule
     \multicolumn{1}{l}{Overall F1} & 
     \multicolumn{15}{c}{M1: 93.03; M2: 92.22} &
     \multicolumn{15}{c}{M1: 45.96; M2: 45.33} &
     \multicolumn{15}{c}{M1: 85.56; M2: 85.70}  &
     \multicolumn{15}{c}{M1: 77.23; M2: 78.71} &
     \multicolumn{15}{c}{M1: 87.92; M2: 89.35}&
     \multicolumn{15}{c}{M1: 63.38; M2: 63.26}\\
    \cmidrule(r){1-1}\cmidrule(lr){2-16}\cmidrule(lr){17-33}\cmidrule(lr){34-49}\cmidrule(lr){49-63}\cmidrule(lr){64-79}\cmidrule(lr){80-95} 

    & \multicolumn{15}{c}{\multirow{12}[2]{*}{\includegraphics[scale=0.39]{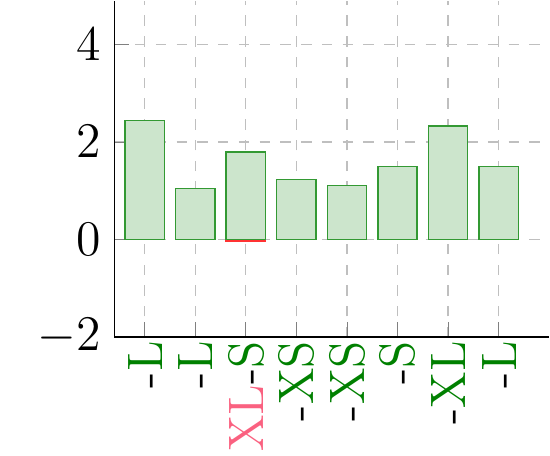}}}              
    & \multicolumn{15}{c}{\multirow{12}[2]{*}{\includegraphics[scale=0.39]{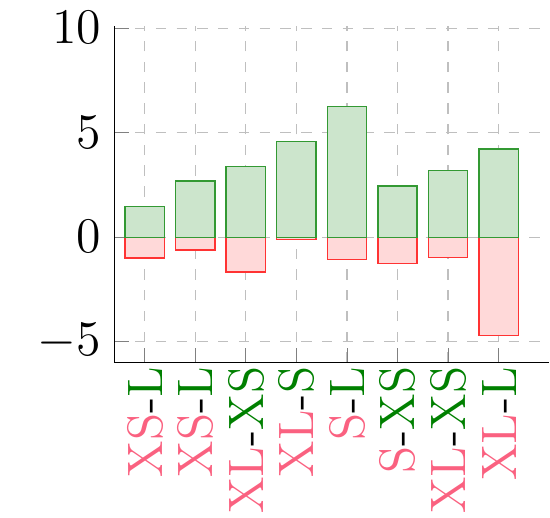}}}              
    & \multicolumn{15}{c}{\multirow{12}[2]{*}{\includegraphics[scale=0.39]{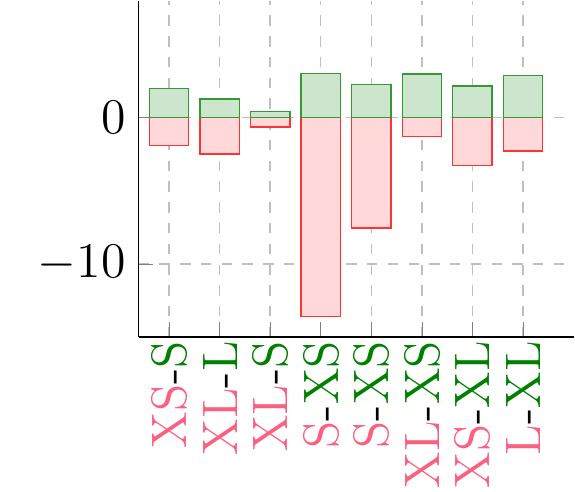}}}              
    & \multicolumn{15}{c}{\multirow{12}[2]{*}{\includegraphics[scale=0.39]{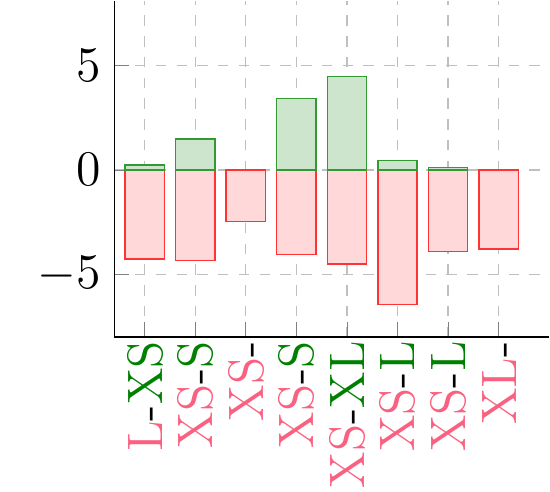}}} 
    &\multicolumn{15}{c}{\multirow{12}[2]{*}{\includegraphics[scale=0.39]{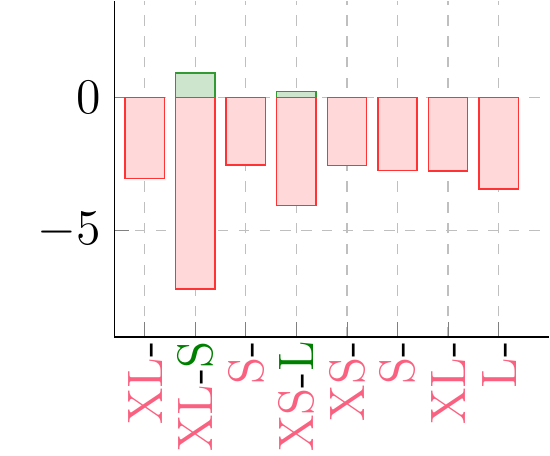}}} 
    &\multicolumn{15}{c}{\multirow{12}[2]{*}{\includegraphics[scale=0.39]{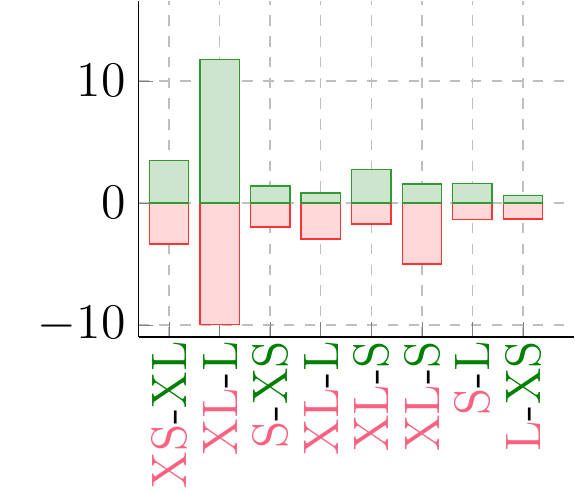}}} 
    \\ \\ \\ \\
     \multicolumn{1}{l}{M1: \textit{CflairWgloveLstmCrf}} &  
    \\ \\ 
    \multicolumn{1}{l}{M2: \textit{CelmWgloveLstmCrf}} & 
    \\  \\
    \multicolumn{1}{c}{\textcolor{brinkpink}{Comparative diagnosis}} & 
    \\ \\ \\ \\ 
    \bottomrule
    \end{tabular}%
    \vspace{-5pt}
  \caption{\textit{Comparative diagnosis} of different NER systems. M1 and M2 denote two models.
   We classify the attribute values into four categories: extra-small (XS), small (S), large (L), and extra-large (XL).
 In the \textit{comparative diagnosis} histogram, \textcolor{aogreen}{green} (\textcolor{brinkpink}{red}) $x$ ticklabels represents the bucket value of a specific attribute on which system M1 surpasses (under-performs) M2 by the largest margin that is illustrated by a \textcolor{aogreen}{green} (\textcolor{brinkpink}{red}) bin. 
 }
  \label{tab:bucket-wise-appendix}%
\end{table*}%

\section{Model-wise Analysis and Observation}
Tab.~\ref{tab:model-wise-conll} gives the model-wise measures $\mathbf{S}^{\rho}_{i,j}$ and $\mathbf{S}^{\sigma}_{i,j}$ which are the average case on all the datasets. We find that: 
\textbf{pre-trained knowledge enhanced models are tardier to the token-level attribute.} We observe that the values of $\mathbf{S}^{\rho}$ dropped sharply on \texttt{tCon} and \texttt{tFre}, when the pre-trained embedding is introduced,   therefore, comparing with the models without pre-trained knowledge, the performance of the models with pre-trained knowledge is slower improved as the increasing of \textit{token consistency} and \textit{token frequency}.
Specifically, the models with pre-trained knowledge have higher performance and lower $\mathbf{S}^{\sigma}$, compared with the models without pre-trained knowledge.
This reveals that the introduction of external knowledge will handle the lower \textit{label consistency} of token and low \textit{token frequency}.

\renewcommand\tabcolsep{2.8pt}
\begin{table*}[!th]
  \centering \footnotesize
    \begin{tabular}{lcccccccc}
    \toprule
    \textbf{dataset} & \textbf{eDen} & \textbf{oDen} & \textbf{sLen} & \textbf{eCon} & \textbf{eFre} & \textbf{tCon} & \textbf{tFre} & \textbf{eLen} \\
    \midrule
    \textbf{conll03} & ${2.2\times10^{-12}}$ & ${2.0\times10^{-17}}$ & ${1.1\times10^{-10}}$ & ${1.0\times10^{-6}}$ & ${1.2\times10^{-14}}$ & ${8.8\times10^{-18}}$ & ${8.2\times10^{-11}}$ & ${4.8\times10^{-7}}$ \\
    \textbf{wnut16} & ${2.6\times10^{-15}}$ & ${7.3\times10^{-17}}$ & ${1.1\times10^{-13}}$ & ${1.4\times10^{-6}}$ & ${4.8\times10^{-10}}$ & ${3.7\times10^{-15}}$ & ${1.4\times10^{-14}}$ & ${4.8\times10^{-7}}$ \\
    \textbf{notewb} & ${3.9\times10^{-16}}$ & ${9.0\times10^{-13}}$ & ${5.5\times10^{-09}}$ & ${7.5\times10^{-8}}$ & ${2.1\times10^{-16}}$ & ${2.1\times10^{-18}}$ & ${8.0\times10^{-17}}$ & ${3.6\times10^{-7}}$ \\
    \textbf{notemz} & ${3.6\times10^{-11}}$ & ${5.1\times10^{-11}}$ & ${2.2\times10^{-11}}$ & ${1.3\times10^{-6}}$ & ${5.3\times10^{-12}}$ & ${2.9\times10^{-18}}$ & ${6.1\times10^{-16}}$ & ${5.5\times10^{-7}}$ \\
    \textbf{notebc} & ${1.7\times10^{-05}}$ & ${3.8\times10^{-11}}$ & ${8.8\times10^{-13}}$ & ${1.3\times10^{-6}}$ & ${6.3\times10^{-15}}$ & ${4.1\times10^{-18}}$ & ${5.2\times10^{-15}}$ & ${5.5\times10^{-7}}$ \\
    \textbf{notebn} & ${2.9\times10^{-07}}$ & ${1.6\times10^{-11}}$ & ${5.7\times10^{-14}}$ & ${1.3\times10^{-7}}$ & ${2.9\times10^{-15}}$ & ${2.9\times10^{-18}}$ & ${2.4\times10^{-15}}$ & ${7.5\times10^{-8}}$ \\
    \bottomrule
    \end{tabular}%
    \caption{$p$-values from the Friedman test. The null hypothesis is that the performance of different buckets with respect to an attribute has the same means for a given \textbf{dataset}.}
  \label{tab:signif-data}%
\end{table*}%

\section{Bucket-wise Analysis and Observation}

Tab.~\ref{tab:bucket-wise-appendix} illustrates the comparative diagnosis of different NER systems. Here,  we will give the observations.

\paragraph{LSTM v.s. CNN}

The sentence encoder of CNN is better at dealing with long entities (\texttt{eLen:XL}) on the datasets with a high value of $\zeta_{eCon}$. 
As shown in Tab.\ref{tab:bucket-wise-appendix}, the performance of LSTM and CNN systems are significantly different on the ``\texttt{eLen:XL}'' bucket ($p=1.2\times10^{-2}<0.05$) without regard to \texttt{WNUT16} and \texttt{WB} two datasets which have the lowest values of  $\zeta_{eCon}$.

The encoder of LSTM does better in dealing with highly-ambiguous entities (\texttt{eCon:S}).
For example, the LSTM system has surpassed CNN on the datasets \texttt{WNUT} and \texttt{WB}, whose average label ambiguities of entities are the two largest ones.

\renewcommand\tabcolsep{3.6pt}
\begin{table*}[!htb]
  \centering \footnotesize
    \begin{tabular}{lcccccccc}
    \toprule
    \textbf{Model} & \textbf{eDen} & \textbf{oDen} & \textbf{sLen} & \textbf{eCon} & \textbf{eFre} & \textbf{tCon} & \textbf{tFre} & \textbf{eLen} \\
    \midrule
    \textit{CRF++} & \cellcolor{mypink}0.39  & \cellcolor{mypink}0.31  & \cellcolor{mypink}0.28  & $9.4\times10^{-4}$ & $1.8\times10^{-3}$ & $9.4\times10^{-4}$ & $9.7\times10^{-3}$ & $3.8\times10^{-3}$ \\
    \textit{CnoneWrandLstmCrf} & \cellcolor{mypink}0.09  & \cellcolor{mypink}0.17  & \cellcolor{mypink}0.10  & $1.0\times10^{-3}$ & $1.0\times10^{-3}$ & $9.4\times10^{-4}$ & $1.8\times10^{-3}$ & $3.8\times10^{-3}$ \\
    \textit{CcnnWnoneLstmCrf} & \cellcolor{mypink}0.10  & \cellcolor{mypink}0.80  & \cellcolor{mypink}0.80  & $3.2\times10^{-3}$ & $9.4\times10^{-4}$ & $1.5\times10^{-3}$ & $2.9\times10^{-2}$ & $5.6\times10^{-3}$ \\
    \textit{CcnnWrandLstmCrf} & \cellcolor{mypink}0.46  & \cellcolor{mypink}0.56  & \cellcolor{mypink}0.85  & $2.2\times10^{-3}$ & $7.1\times10^{-4}$ & $9.4\times10^{-4}$ & $1.8\times10^{-3}$ & $3.8\times10^{-3}$ \\
    \textit{CcnnWgloveLstmCrf} & \cellcolor{mypink}0.61  & \cellcolor{mypink}0.28  & \cellcolor{mypink}0.49  & $1.5\times10^{-3}$ & $5.6\times10^{-3}$ & $1.1\times10^{-3}$ & $9.7\times10^{-3}$ & $1.5\times10^{-3}$ \\
    \textit{CcnnWgloveCnnCrf} & \cellcolor{mypink}0.61  & \cellcolor{mypink}0.39  & \cellcolor{mypink}0.80  & $1.5\times10^{-3}$ & $6.3\times10^{-3}$ & $1.7\times10^{-3}$ & $1.5\times10^{-3}$ & $2.0\times10^{-3}$ \\
    \textit{CcnnWgloveLstmMlp} & \cellcolor{mypink}0.39  & \cellcolor{mypink}0.46  & \cellcolor{mypink}0.33  & $1.1\times10^{-3}$ & $2.9\times10^{-3}$ & $1.1\times10^{-3}$ & $5.6\times10^{-3}$ & $6.7\times10^{-3}$ \\
    \textit{CelmoWnoneLstmCrf} & \cellcolor{mypink}0.26  & \cellcolor{mypink}0.57  & \cellcolor{mypink}0.33  & $2.0\times10^{-3}$ & $4.2\times10^{-3}$ & $1.1\times10^{-3}$ & $5.6\times10^{-3}$ & $5.6\times10^{-3}$ \\
    \textit{CelmoWgloveLstmCrf} & \cellcolor{mypink}0.85  & \cellcolor{mypink}0.10  & \cellcolor{mypink}0.22  & $2.0\times10^{-3}$ & $3.8\times10^{-3}$ & $1.1\times10^{-3}$ & $8.1\times10^{-3}$ & $1.5\times10^{-3}$ \\
    \textit{CbertWnonLstmMlp} & \cellcolor{mypink}0.06  & \cellcolor{mypink}0.12  & \cellcolor{mypink}0.61  & $3.8\times10^{-3}$ & $4.2\times10^{-3}$ & $2.0\times10^{-3}$ & $2.0\times10^{-2}$ & $3.5\times10^{-2}$ \\
    \textit{CflairWnoneLstmCrf} & \cellcolor{mypink}0.13  & \cellcolor{mypink}0.22  & \cellcolor{mypink}0.39  & $2.2\times10^{-3}$ & $3.8\times10^{-3}$ & $1.1\times10^{-3}$ & $3.8\times10^{-3}$ & $5.6\times10^{-3}$ \\
    \textit{CflairWgloveLstmCrf} & \cellcolor{mypink}0.39  & \cellcolor{mypink}0.33  & \cellcolor{mypink}0.20  & $4.6\times10^{-3}$ & $2.2\times10^{-3}$ & $1.1\times10^{-3}$ & \cellcolor{mypink}$6.0\times10^{-2}$ & $3.8\times10^{-3}$ \\
    \bottomrule
    \end{tabular}%
      \caption{$p$-values from the Friedman test. The null hypothesis is that the performance of different buckets with respect to an attribute has the same means for a given \textbf{model}. The \texttt{Pink} region denote the attribute on the given model does not pass ($p\geq0.05$) a significance test at $p=0.05$. }
  \label{tab:signif-model}%
\end{table*}%

\paragraph{Flair v.s. ELMo}
\label{sec:comp-diagnosis-flair}
While the current state-of-the-art NER model (\texttt{Flair}) has achieved the best performance in terms of dataset-level F1 score, a worse-ranked model (\texttt{ELMo}) can outperform it in some attributes.
Typically, Flair performs worse when dealing with long sentences, which holds for all the datasets ($p=1.4\times10^{-3}<0.05$). The reason can be attributed to its structural bias, which adopts an LSTM-based encoder for character language modeling, suffering from long-term dependency problems. One potential promising improvement is resorting to the Transformer-based architecture for the character language model pre-training.

\section{Significance Testing}
We break down the holistic performance into different categories for conducting the fine-grained evaluation. Specifically, we divide the set of test entities (or tokens) into different subsets (we named buckets) of test entities. To test whether the performance of buckets with respect to an attribute is significantly different, we perform Friedman significance testing at $p=0.05$ in dataset-dimension and model-dimension.  To ensure a sufficient sample size to conduct significance testing, we restarted a model on the same dataset for twice. 

\textbf{Dataset-dimension significance testing} It is the premise of attribute-wise analysis. The null hypothesis is that the performance of different buckets with respect to an attribute has the same means for a given \textbf{dataset}. The significance testing results are shown in Tab.~\ref{tab:signif-data}. The $p$-values of these eight attributes on the six datasets are smaller than $0.05$, indicating that the performance of buckets with respect to one of the eight attributes is significantly different for a given dataset.   

\textbf{Model-dimension significance testing} It is the premise of model-wise analysis. The null hypothesis is that the performance of different buckets with respect to an attribute has the same means for a given \textbf{model}. The significance testing results are shown in Tab.~\ref{tab:signif-model}. 
We observe that the $p$-values of \texttt{eDen}, \texttt{oDen}, and \texttt{sLen} are larger than $0.05$, therefore, \texttt{eDen}, \texttt{oDen}, and \texttt{sLen} does not pass the significance testing for a given model. The performance of the buckets with respect to  \texttt{eDen} (\texttt{oDen},  \texttt{sLen}) are not significantly different.

\end{document}